\documentclass[10pt]{article}

\pdfoutput=1

\usepackage{microtype}
\usepackage[margin=1in]{geometry}
\usepackage{hyperref}

\usepackage{xcolor}
\usepackage{graphicx, subcaption, caption, stackengine}
\usepackage{mathtools, amsmath, amssymb, amsfonts, amsthm}
\usepackage{enumitem} 
\usepackage{algorithm, algorithmic}
\usepackage{todonotes}
\usepackage{hyperref}
\usepackage{array}
\usepackage{stfloats}
\usepackage{siunitx}

\newtheorem{theorem}{Theorem}[section]

\newtheorem{prop}[theorem]{Proposition}
\newtheorem{corollary}{Corollary}[theorem]

\theoremstyle{definition}
\newtheorem{definition}{Definition}[section]

\DeclareMathOperator*{\argmax}{arg\,max}
\DeclareMathOperator*{\argmin}{arg\,min}
\DeclarePairedDelimiter{\norm}{\lVert}{\rVert}
\DeclarePairedDelimiter{\abs}{\lvert}{\rvert}
\DeclarePairedDelimiterX{\ip}[2]{\langle}{\rangle}{#1, #2}
\DeclareMathOperator{\E}{\mathbb{E}}

\DeclareMathOperator{\A}{A}
\DeclareMathOperator{\med}{med}

\newcommand{\R}{\mathbb{R}}
\newcommand{\N}{\mathbb{N}}
\newcommand{\X}{\mathcal{X}}
\newcommand{\Ucal}{\mathcal{U}}
\newcommand{\C}{\mathcal{C}}
\newcommand{\Lcal}{\mathcal{L}}

\graphicspath{{./figures/}{./figures/system/}{./figures/illustrative/}{./figures/results/}{./figures/failure/}}

\usepackage{wrapfig}
\usepackage[round]{natbib}
\usepackage{placeins}
\newcolumntype{M}[1]{>{\centering\arraybackslash}m{#1}}
\newcolumntype{L}[1]{>{\raggedright\arraybackslash}m{#1}}

\title{Feedback Coding for Active Learning}

\author{Gregory Canal\qquad
Matthieu Bloch\qquad
Christopher Rozell\\
\\
School of Electrical and Computer Engineering\\
Georgia Institute of Technology, Atlanta, GA\\
\{gregory.canal, matthieu, crozell\}@gatech.edu}

\begin{document}

\date{}
\maketitle
 
\begin{abstract}
The iterative selection of examples for labeling in active machine learning is conceptually similar to feedback channel coding in information theory: in both tasks, the objective is to seek a minimal sequence of actions to encode information in the presence of noise. While this high-level overlap has been previously noted, there remain open questions on how to best formulate active learning as a communications system to leverage existing analysis and algorithms in feedback coding. In this work, we formally identify and leverage the structural commonalities between the two problems, including the characterization of encoder and noisy channel components, to design a new algorithm. Specifically, we develop an optimal transport-based feedback coding scheme called \emph{Approximate Posterior Matching} (APM) for the task of active example selection and explore its application to Bayesian logistic regression, a popular model in active learning. We evaluate APM on a variety of datasets and demonstrate learning performance comparable to existing active learning methods, at a reduced computational cost. These results demonstrate the potential of directly deploying concepts from feedback channel coding to design efficient active learning strategies.
\end{abstract}

\section{Introduction}
Active learning is an area of modern machine learning that studies how data points can be sequentially selected for labeling to train a model with as few labeled examples as possible \citep{settles2009active}. Minimizing the number of labeled examples is critical in any learning scenario where labels are expensive to obtain, such as in healthcare applications where a medical expert must hand-label each training example \citep{liu2004}, or where only a limited number of examples can be evaluated, such as in drug discovery \citep{warmuth2003}.

The active selection of data points shares many technical parallels with channel coding with feedback, where a message is encoded into a sequence of symbols transmitted across a noisy channel and each symbol is selected based on the message and past channel outputs. In active learning, the optimal classifier parameters play the role of the ``message'' while the sequence of examples with noisy labels plays the role of ``channel outputs'' available through feedback to select the next example for labeling. Both feedback channel coding and active learning seek to minimize the number of encoder actions, leverage a history of noisy observations to select the next most informative action, must account for observation noise, and should operate in a computationally efficient manner. Although there exists a large literature studying the intersection of information theory with machine learning \citep{xu2017} and specifically active learning \citep{naghshvar2015}, there remain open questions about the best ways to directly leverage techniques in channel coding for active example selection.

The main contribution of this work is a formulation of general active learning problems in terms of a feedback coding system, and a demonstration of this approach through the application and analysis of active learning in logistic regression. To motivate this approach, we first examine active learning through the lens of feedback channel coding by identifying communications system components, including a deterministic encoder, noisy channel, channel input constraints, and capacity-achieving distribution. With these components identified, we show how typical structural constraints in active learning problems prevent the direct application of existing feedback coding approaches such as \emph{posterior matching} \citep{ma2011}. We address this challenge by proposing \emph{Approximate Posterior Matching} (APM), an optimal transport-based active learning scheme that extends posterior matching to account for the type of encoder constraints found in active learning problems.

To demonstrate the power of this approach, we apply APM to Bayesian logistic regression, a popular model in active learning. We identify the communication system components in logistic regression, derive a corresponding APM selection scheme (APM-LR), provide analytical results concerning each selected example's information content, and empirically demonstrate on several datasets how APM-LR attains a sample complexity comparable to other active logistic regression methods at a reduced computational cost. While this example scenario highlights the capabilities of APM as a specific data selection method, the feedback communications framework we develop provides a unified approach for designing and analyzing active learning systems in general.
\subsection{Related Work}
Modern active learning methods vary considerably in their approach to example selection, ranging from coreset construction \citep{pinsler2019,sener2018active} and adversarial learning of informative examples \citep{Sinha_2019_ICCV} to ensemble measures of example utility \citep{Beluch_2018_CVPR} and Bayesian information acquisition methods \citep{gal2017, kirsch2019}. Bayesian active learning methods are intimately related to concepts in information and coding theory, and the intersection between these topics has a long history rooted in the study of sequential design of experiments \citep{lindley1956, chernoff1959} and active hypothesis testing \citep{burnashev1974}. Since this early work, direct estimation and maximization of information gain has emerged as a popular active learning method \citep{mackay1992}, and has been approximated for computational tractability \citep{houlsby2011bayesian}. More recently, \citet{naghshvar2015} have studied the direct application of an information-theoretic active hypothesis testing method to active learning problems. This method is limited to discriminating between a finite number of hypotheses (as opposed to estimating arbitrary model parameters) and to our knowledge has not been applied to popular machine learning models such as logistic regression. Other works have described at a high-level the similarities between active learning and coding with feedback over a noisy channel but do not exploit this observation to leverage existing coding schemes for example selection \citep{chen2015, castro2013}.

Posterior matching \citep{shayevitz2011, ma2011} is a general feedback coding scheme that has been applied to tasks beyond telecommunications such as brain-computer interfacing \citep{omar2010, tantiongloc2017} and aircraft path planning \citep{akce2010}, but has limited application to example selection in active learning. \citet{castro2008} study an active learning algorithm related to posterior matching that learns decision boundaries in discretized spaces, but does not directly maximize information about hyperplane parameters in a continuous space as we do here. More generally, to our knowledge existing work has not framed the task of active learning as a feedback communications system for the purpose of identifying an equivalent capacity-achieving distribution and selecting examples whose channel input distribution most closely approximates it, as we do here.

Logistic regression is a popular setting for the study of active learning, and has served as a testbed for the evaluation of competing example selection techniques. \citet{yang2018} surveyed modern active learning methods for logistic regression and evaluated them on many datasets. They generally found that uncertainty sampling and random sampling match or exceed the performance of more sophisticated (and computationally intensive) example selection methods. Uncertainty sampling, where examples closest to the estimated decision boundary are selected for labeling, is arguably the most popular active learning method for linear classification \citep{tong2001support}. Other active learning methods for linear classifiers are discussed in the literature related to learning halfspaces under bounded noise \citep{zhang2020efficient}. 
\section{Active Learning as a Communications Model}
Let $\Ucal \subseteq \R^d$ denote a pool of unlabeled examples from which at each training iteration $n \in \N$ an example $x_n \in \Ucal$ is selected for labeling by an expert, who assigns label $Y_n \in \{1,2,\dots K\}$ according to a probabilistic model (all random variables are capitalized in this work). We consider a Bayesian framework in which we assume the existence of ground-truth model parameters $\theta \in \Theta$ distributed according to a prior $p_\theta$ that parameterizes a distribution $p(Y \mid x,\theta)$ governing the expert's labeling behavior. As is common in active learning, we assume that the labels $\{Y_n\}$ are independent when conditioned on $\theta$. At each iteration $n$, a learning algorithm $\A$ is trained on a labeled dataset $\Lcal_n = \{(x_i,y_i)\}_{i=1}^n$ (using lowercase to denote previously observed labels), resulting in a trained model with parameters $\widehat{\theta}_n \in \Theta$. The task of active learning is to design a policy $\pi_n$ that, at each iteration, uses the label history $\Lcal_{n-1}$ to select example $x_n$ from the remaining unlabeled examples $\Ucal_n \coloneqq \Ucal \setminus \{x_i\}_{i=1}^{n-1}$, such that the classifier trains a generalizable model with as few labeled examples as possible. 

\begin{wrapfigure}{R}{0.5\textwidth}
	\def\cw{0.1}
	\def\fw{0.85}
	\def\vh{5mm}
	\centering
	\begin{subfigure}[t]{0.5\textwidth}
		\centering
		\parbox{\cw\textwidth}{\caption{}\label{fig:LR-as-coding}}
		\parbox{\fw\textwidth}{\includegraphics[width=\linewidth]{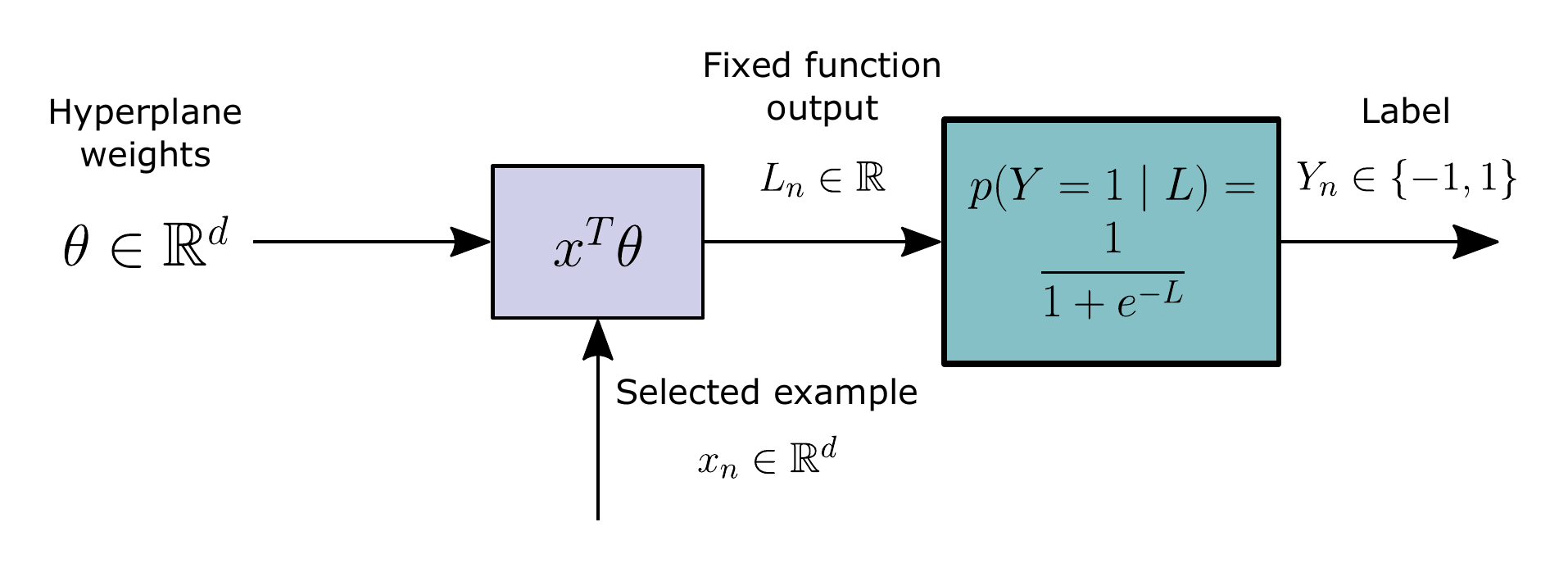}}
	\end{subfigure}%
	\\
	\vspace{\vh}
	\begin{subfigure}[t]{0.5\textwidth}
		\centering
		\parbox{\cw\textwidth}{\caption{}\label{fig:AL-as-coding}}
		\parbox{\fw\textwidth}{\includegraphics[width=\linewidth]{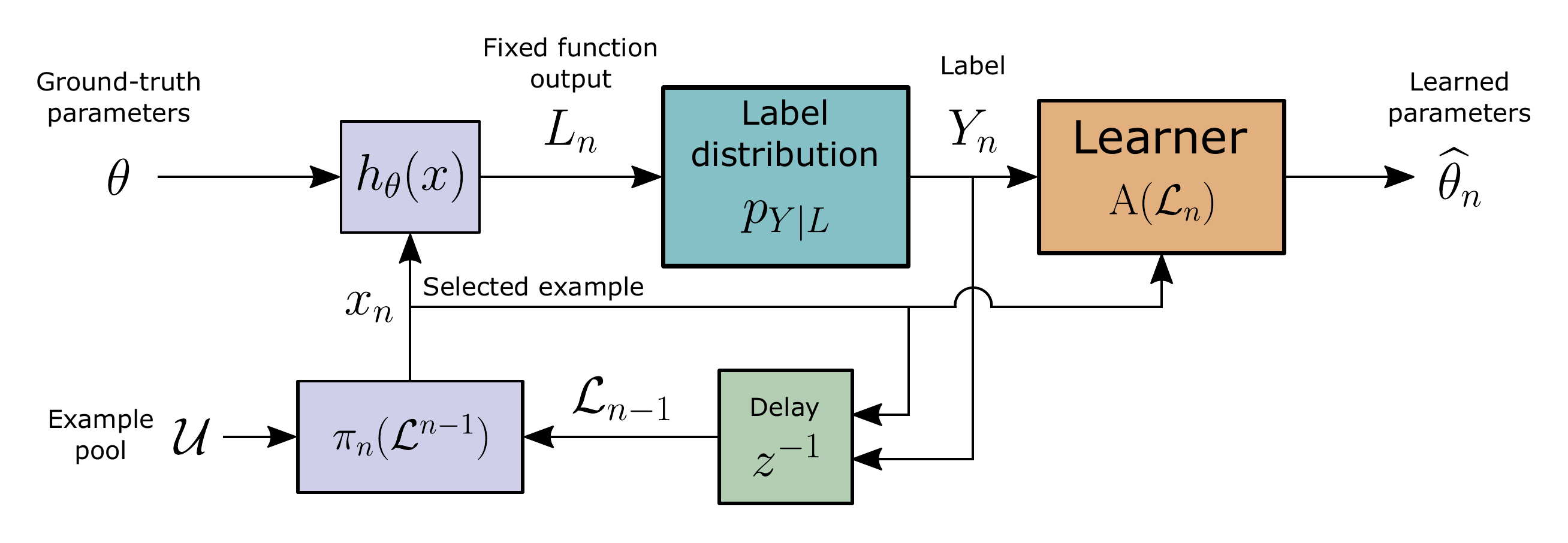}}
	\end{subfigure}%
	\\
	\vspace{\vh}
	\begin{subfigure}[t]{0.5\textwidth}
		\centering
		\parbox{\cw\textwidth}{\caption{}\label{fig:feedback-coding}}
		\parbox{\fw\textwidth}{\includegraphics[width=\linewidth]{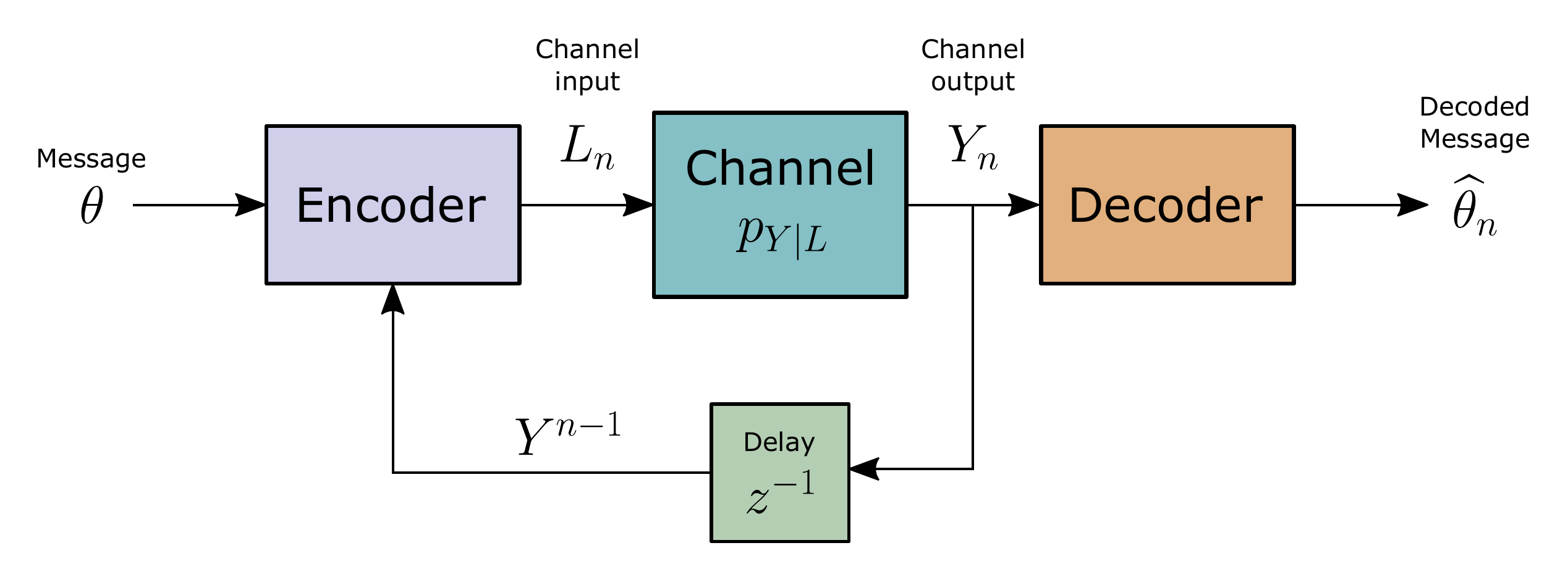}}
	\end{subfigure}%
	\caption{(a) Decomposition of logistic regression into an inner product between hyperplane $\theta$ and example $x_n$, and a logistic label distribution that depends only on this product. (b) Active learning decomposed into a deterministic function $h$, label distribution $p(Y \mid L)$, and feedback of the labeling history $\Lcal_{n-1}$ to example selection policy $\pi_n$. (c) Coding with feedback, where a message is transmitted across a noisy channel as a sequence of symbols and subsequently decoded. By comparing (b) and (c), one can draw direct connections between active learning and coding with feedback.}
	\label{fig:info-as-comms}
\end{wrapfigure}
In active logistic regression, $\theta$ encodes the weights of a linear separator, with $\Theta = \R^d$ (we consider only homogeneous logistic regression in this work). We assume a Gaussian prior $p_\theta \sim \mathcal{N}(0, \frac1\lambda I)$ with hyperparameter $\lambda > 0$. The label $Y \in \{-1,1\}$ for data example $x$ is assumed to be distributed according to
\begin{equation}
p(Y=1 \mid x, \theta) = \frac{1}{1+e^{-x^T \theta}}\label{eq:LR-likelihood}.
\end{equation}
Given a labeled dataset $\Lcal$, we consider a maximum a posteriori (MAP) learning algorithm given by the convex program
{\small
	\begin{align}
	\A(\Lcal) &= \argmax_{\theta \in \R^d} \enspace \ln \, p_\theta \!\prod_{(x,y) \in \Lcal} p(y \mid x,\theta)\notag
	\\ &= \argmin_{\theta \in \R^d} \enspace \frac{\lambda}{2} \norm{\theta}_2^2 \, + \! \sum_{(x,y) \in \Lcal} \ln(1 + e^{-y x^T \theta}).\label{eq:LR-MAP}
	\end{align}}
Our key insight in this work is to define an intermediate variable $L=h_\theta(x)$, where $h_\theta(x) \coloneqq x^T \theta$, and decompose the labeling distribution in \eqref{eq:LR-likelihood} into $p(Y=1\mid x, \theta) = p(Y=1\mid L) = \frac{1}{1 + e^{-L}}$. This decomposition of the labeling distribution into a deterministic function $h_\theta(x)$ and conditional distribution $p(Y \mid L)$ can be found in many machine learning models. For instance, in Bayesian neural networks \citep{gal2017}, $h_\theta(x)$ is typically given by the composition of several nonlinear layers with $L=h_\theta(x)$ encoding the final layer feature vector, and $p(Y \mid L)$ is given by the softmax function. Figure \ref{fig:LR-as-coding} depicts this decomposition for logistic regression, and Figure \ref{fig:AL-as-coding} illustrates the full active learning decomposition in the general case.

By decomposing active learning in this manner, we are able to draw direct connections to feedback channel coding, in which a message $\theta$ is encoded into a sequence of symbols $\{L_n\}$, transmitted across a channel with transition probability $p(Y \mid L)$ yielding noisy output symbols $\{Y_n\}$, and subsequently decoded into an estimated message $\widehat{\theta}_n$. The availability of noiseless feedback from the channel output to the encoder provides the encoder with the history of received symbols, and allows it to adaptively select an informative channel input (Figure \ref{fig:feedback-coding}). By comparing Figures \ref{fig:AL-as-coding} and \ref{fig:feedback-coding}, we can see the direct correspondence between active learning and channel coding with feedback: model parameters $\theta$ serve as the message, which is encoded by function $h$ (parameterized by $x_n$) into channel input $L_n = h_\theta(x_n)$. Label distribution $p(Y \mid L)$ can be interpreted as a noisy channel, with label $Y_n$ as the channel output. Algorithm $\A$ decodes labeled data $\Lcal_n$ into a decoded message $\widehat{\theta}_n$, and $\Lcal_n$ is passed as noiseless feedback to the encoder. This formulation of active learning as a feedback communications system allows one to leverage existing tools in channel coding for the design of an example selection scheme $\pi_n$. While similar decompositions have been observed in prior work \citep{naghshvar2015,chen2015}, we believe our work is the first to use this approach to analyze active learning in a real-world setting such as logistic regression.

\subsection{Optimal Feedback Coding}

In devising a feedback coding scheme for selecting a sequence of channel inputs $\{L_n\}$, there are several quantities that characterize optimal performance. We denote the \emph{mutual information} $I(L;Y)$ between random variables $L$ and $Y$ as a function of marginal distribution $p_L$ and conditional distribution $p_{Y \mid L}$ (using the notation $p_L$ and $p_{Y \mid L}$ interchangeably with $p(L)$ and $p(Y \mid L)$) given by $I(p_L,p_{Y \mid L})$:
\[I(p_L,p_{Y \mid L}) \coloneqq \int_{L,Y} p_L\,p_{Y \mid L} \log_2\frac{p_{Y \mid L}}{p_Y},\]
where $p_Y$ denotes the output distribution of channel $p_{Y \mid L}$ with input distribution $p_L$. Letting $y^i \coloneqq \{y_1,\dots y_{i}\}$ denote the history of observed channel outputs, at iteration $n$ we seek to maximize the \emph{information gain} $I(\theta;Y_n \mid y^{n-1})$, which measures the one-step decrease in uncertainty about the message upon receiving each channel output. For deterministic encoders, information gain is equal to $I(L_n ; Y_n \mid y^{n-1})= I(p_{L_n \mid y^{n-1}}, p_{Y \mid L})$ \citep{cover2006}. Note that for a fixed channel $p_{Y \mid L}$, information gain is only a function of the channel input distribution $p_{L_n \mid y^{n-1}}$, conditioned on the history of channel outputs.

A key quantity in channel coding is the \emph{channel capacity} $C$, defined as the maximum mutual information across the channel for any channel input distribution $p_L$ within some class $\C$:
\[p^*_L(\C) \coloneqq \argmax_{p_L \in \C}\,I(p_L, p_{Y \mid L})\quad C\coloneqq I(p^*_L,p_{Y \mid L}).\]
The \emph{capacity-achieving distribution} $p^*_L(\C)$ is the input distribution in $\C$ that maximizes information across the channel. Through achievability and converse arguments, a central result in information theory is that optimal coding schemes, when marginalized over the message set, should induce the capacity-achieving distribution on the channel input \citep{shannon1948}. In working towards applying existing feedback coding schemes to active example selection, we first characterize the capacity-achieving distribution for logistic regression, which is a core contribution of our work and forms the basis of our novel active logistic regression scheme in Section \ref{sec:LR}.

\paragraph{Channel Capacity in Logistic Regression.} Letting $f(\ell) \coloneqq \frac{1}{1+e^{-\ell}}$, we observe from Figure \ref{fig:LR-as-coding} that logistic regression has a binary output channel with transition probability $p(Y=1 \mid L) = f(L)$. Without constraints on the channel input, the information gain can be maximized by placing masses of equal weight at $\pm\infty$. However, logistic regression imposes the structural constraint $L = x^T \theta$, so that such a distribution would require data points of infinite energy for finite model weights. Therefore, to characterize logistic regression capacity in practice, we consider the capacity-achieving distribution within the class of power-constrained distributions given by $\C_P \coloneqq \{p_L:\E[L^2] \le P\}$; we discuss the selection of $P$ in Section \ref{sec:LR}. With this class defined, we have our first result.
\begin{prop}[Capacity of Logistic Regression]
	\label{prop:BPSK}
	For $p(Y=1\mid L) = f(L)$, we have $p^*_L(\C_P) = B_{\sqrt{P}}$, where $B_t$ is defined as $B_t(\ell) \coloneqq \frac12 \delta(\ell - t) + \frac12 \delta(\ell + t)$
	and $\delta$ denotes the Dirac delta function. Furthermore, we have $C = I(B_{\sqrt{P}},f) = 1-h_b(f(\sqrt{P}))$, where $h_b$ denotes the binary entropy function.
\end{prop}
The proof follows closely to that of \citet{singh2009} for the one-bit quantized Gaussian channel; the proofs of Proposition \ref{prop:BPSK} and all subsequent results are presented in Appendix \ref{app:proofs}.
\subsection{Posterior Matching}
By characterizing the channel capacity and capacity-achieving distribution of active learning models, we enable the use of existing feedback coding schemes that achieve capacity. Recently, a capacity-achieving feedback coding scheme known as \emph{posterior matching} has been developed to select a sequence of channel inputs $\{L_n\}$ to maximize the information gain across a given channel $p_{Y \mid L}$. The central concept is to construct an encoder that by definition induces $p_{L_n \mid y^{n-1}} = p^*_L$ for every $n$, which in essence hands the decoder the information that it is still ``missing'' \citep{ma2011}. This involves the construction of an encoder mapping $S_{y^{n-1}}\colon \theta \to L$ parameterized by ${y^{n-1}}$ such that $S_{y^{n-1}}(\theta) \sim p^*_L$ for every $n$.

While posterior matching is an attractive feedback coding scheme, there are challenges in applying it to active learning: given the structural constraints of any particular active learning problem as depicted in Figure \ref{fig:AL-as-coding}, it may not always be the case that a mapping from $p_{\theta \mid \Lcal_{n-1}}$ to $p^*_L$ exists, since the encoder is constrained to the set of mappings given by $\{h_\theta(x) : x \in \Ucal_{n}\}$.\footnote{The analogous distribution to $p_{L_n \mid y^{n-1}}$ in active learning is $p_{L_n \mid \Lcal_{n-1}}$. When considering only deterministic example selection schemes, $p_{L_n \mid \Lcal_{n-1}}$ is induced directly from $p_{\theta \mid \Lcal_{n-1}}$, through $h_\theta(x)$.} For example, in active logistic regression under mild assumptions, there exists no $x$ such that $h_\theta(x) \sim p^*_L$, as shown in the following proposition.
\begin{prop}
	\label{prop:BPSK-infeasible-LR}
	Under a log-concave prior distribution $p_\theta$, in Bayesian logistic regression for any $n$ there exists no $x_n$ that induces $p_{L_n \mid \Lcal_{n-1}} \sim p^*_L$.
\end{prop}
Since we assume a Gaussian prior $p_\theta$ (which is log-concave), Proposition \ref{prop:BPSK-infeasible-LR} applies and therefore there exists no active logistic regression scheme $\pi_n$ corresponding to a posterior matching mapping from $\theta$ to $p^*_L$. We suspect that the infeasibility of $p^*_L$ holds generally in other real-world machine learning models (e.g., Bayesian neural networks) due to similar structural constraints imposed by $h_\theta(x)$, preventing the direct application of posterior matching for example selection. In the next section, we extend concepts from posterior matching to a novel active learning scheme compatible with this constrained encoder structure.

\subsection{Approximate Posterior Matching}
To address the impossibility of finding $x \in \Ucal$ that induces $p^*_L$ on $L$, we introduce a scheme that instead selects an example $x_n$ such that $p_{L \mid \Lcal_{n-1}}$ is distributed ``as close as possible'' to $p^*_L$, as measured by a distance between distributions. Specifically, we use the 2-Wasserstein distance because of its convenient geometric properties and compatibility with non-overlapping distribution supports \citep{arjovsky2017wasserstein}. The $p$-Wasserstein distance between distributions $\mu$ and $\nu$ is given by
\[W_p(\mu,\nu) = \biggl(\inf_{\gamma \in \Pi(\mu,\nu)} \int_u \int_v \abs{u-v}^p \gamma(u,v) \biggr)^{\frac1p},\]
where $\Pi(\mu,\nu)$ is the set of couplings with marginal distributions $\mu$ and $\nu$ \citep{villani2008optimal}. Our selection scheme, called $\emph{Approximate Posterior Matching}$ (APM), is then given by
\begin{equation}\label{eq:APM}
x_n = \pi_n(\Lcal_{n-1}) \coloneqq \argmin_{x \in \Ucal_n}\enspace W_2(p_{L_n \mid \Lcal_{n-1}}, p^*_L).
\end{equation}
While APM is intuitively appealing because it steers the induced channel distribution as close as possible to $p^*_L$, we justify this strategy in the next section for the case of logistic regression by showing that information gain does in fact approach its maximum possible value as $W_2(p_{L \mid \Lcal_{n-1}}, p^*_L)$ is minimized.

\section{APM in Logistic Regression}\label{sec:LR}
Under the power constraint $\E[L^2] \le P$, Proposition \ref{prop:BPSK} establishes that the capacity-achieving distribution in the logistic regression system is given by $B_{\sqrt{P}}$. We now show an information continuity result for this capacity-achieving distribution, which provides a mathematical justification for the APM Wasserstein distance minimization in \eqref{eq:APM}.
\begin{theorem}
	\label{thm:infocont} Let $\widetilde{C}_n = \max_{x \in \Ucal_n} I(p_{L_n \mid \Lcal_{n-1}}, f)$ denote the maximum information gain from \emph{any} example selected at iteration $n$, and suppose $P > 0$ is selected such that $p_{L_n \mid \Lcal_{n-1}} \in \C_P$ for any $x \in \Ucal_n$. Then for any $x \in \Ucal_n$,
	\[\widetilde{C}_n - I(p_{L_n \mid \Lcal_{n-1}}, f) \le K_P W_2(p_{L_n \mid \Lcal_{n-1}},B_{\sqrt{P}}),\]
	where $K_P > 0$ is a constant that only depends on $P$.
\end{theorem}
For decreasing $W_2(p_L,B_{\sqrt{P}})$, this result bounds $I(p_{L_n \mid \Lcal_{n-1}}, f)$ towards its maximum possible information gain $\widetilde{C}_n$. In other words, minimizing the distance to the \emph{known} capacity-achieving distribution (even if not achievable in practice) ensures that the information gain approaches its maximum value within the set of possible input distributions --- a value which is \emph{unknown} a priori. As we shall see in the results and experiments that follow, targeting the known capacity-achieving distribution affords geometric simplifications and computational benefits over the strategy of directly selecting the example that achieves $\widetilde{C}_n$. Unlike APM, the latter method does not benefit from analytical knowledge of the information structure of the channel and constraint set, and so it must instead conduct an expensive brute-force maximization of information gain.

\begin{figure*}[t]
	\def\fw{0.25}
	\def\iw{0.9}
	\centering
	\begin{subfigure}[t]{\fw\textwidth}
		\centering
		\stackunder[2pt]{\includegraphics[width=\iw\textwidth]{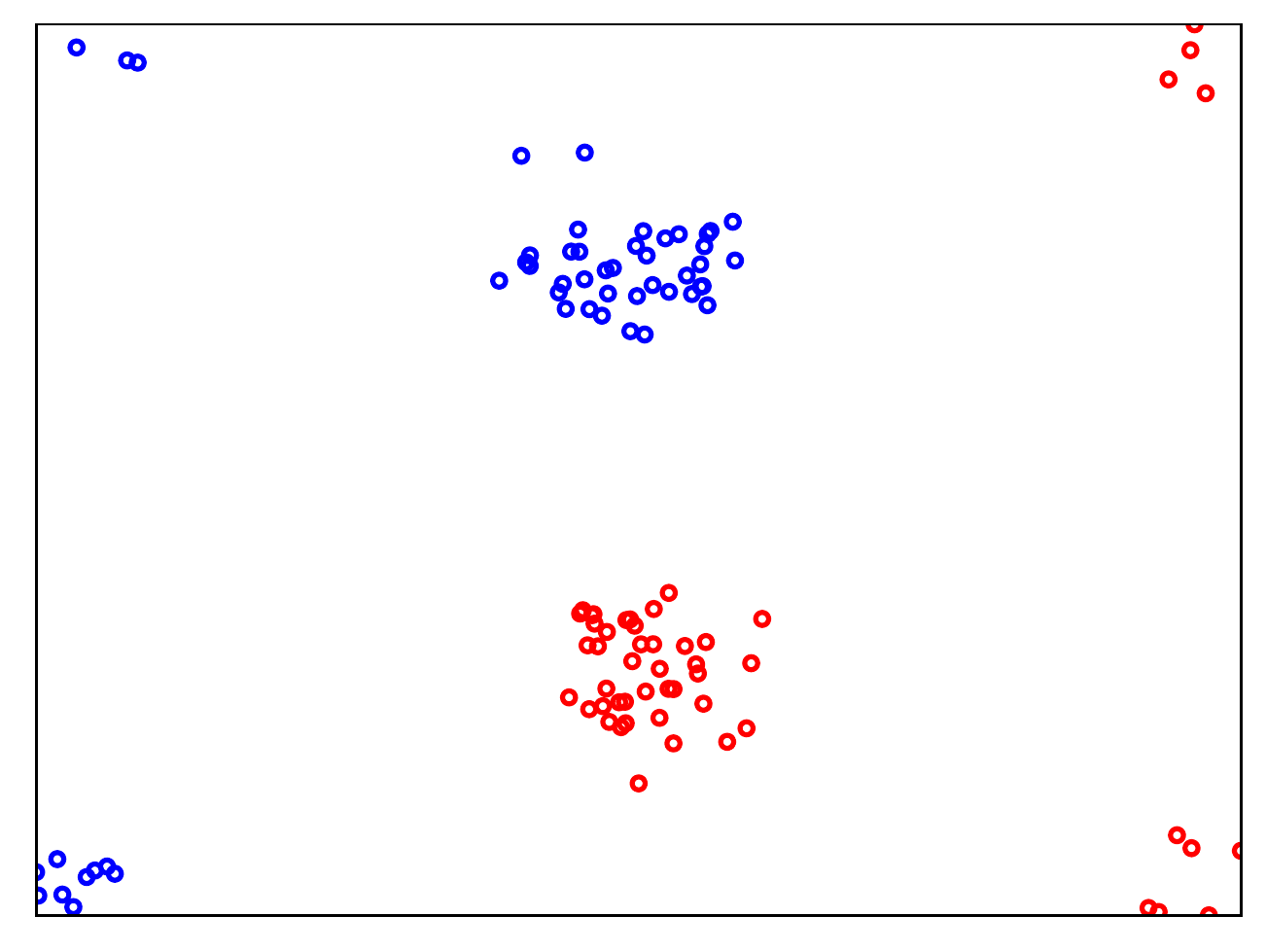}}{\small Labeled data pool}
	\end{subfigure}%
	~
	\begin{subfigure}[t]{\fw\textwidth}
		\centering
		\stackunder[2pt]{\includegraphics[width=\iw\textwidth]{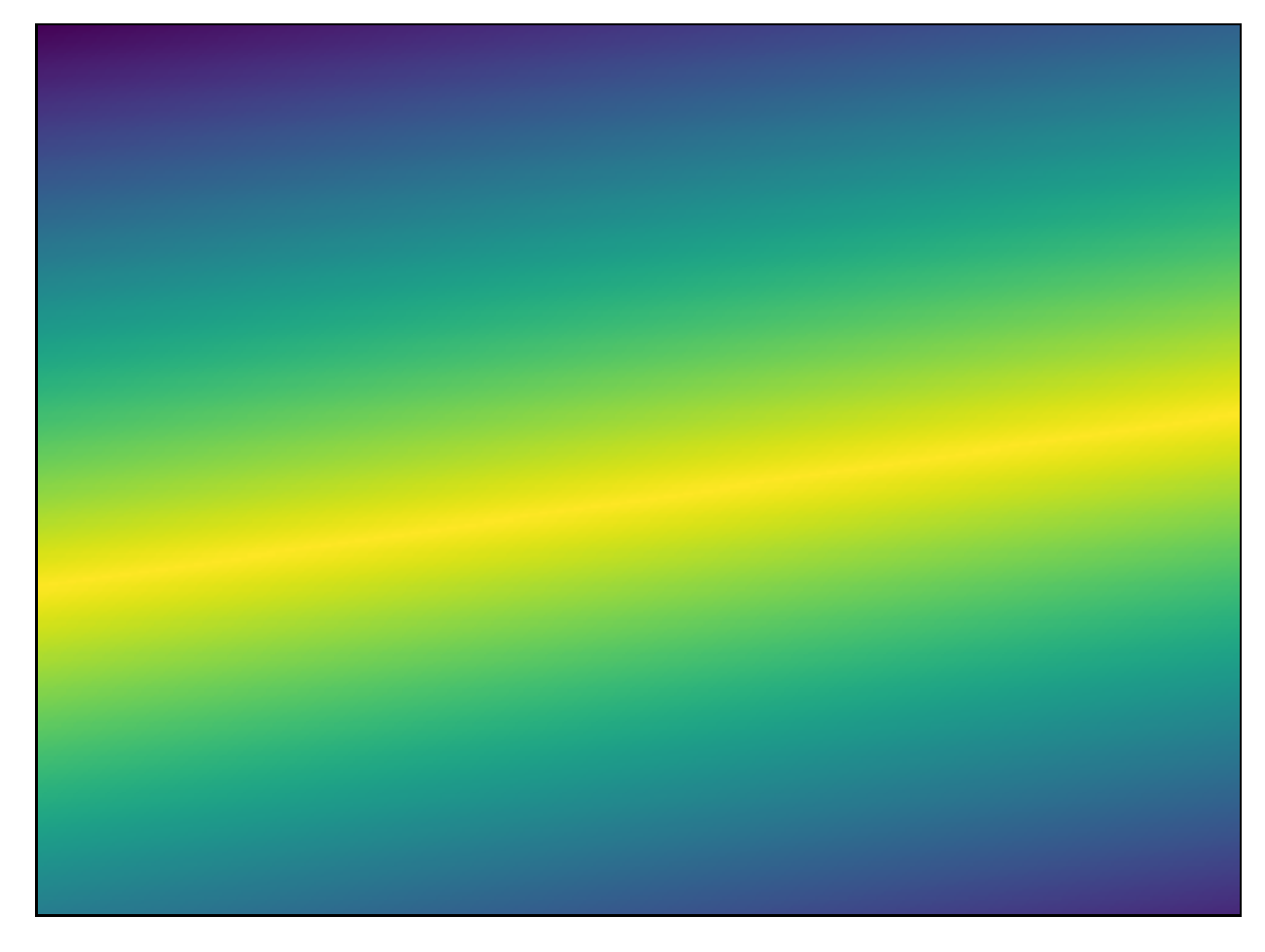}}{\small Uncertainty utility: initial}
	\end{subfigure}%
	~
	\begin{subfigure}[t]{\fw\textwidth}
		\centering
		\stackunder[2pt]{\includegraphics[width=\iw\textwidth]{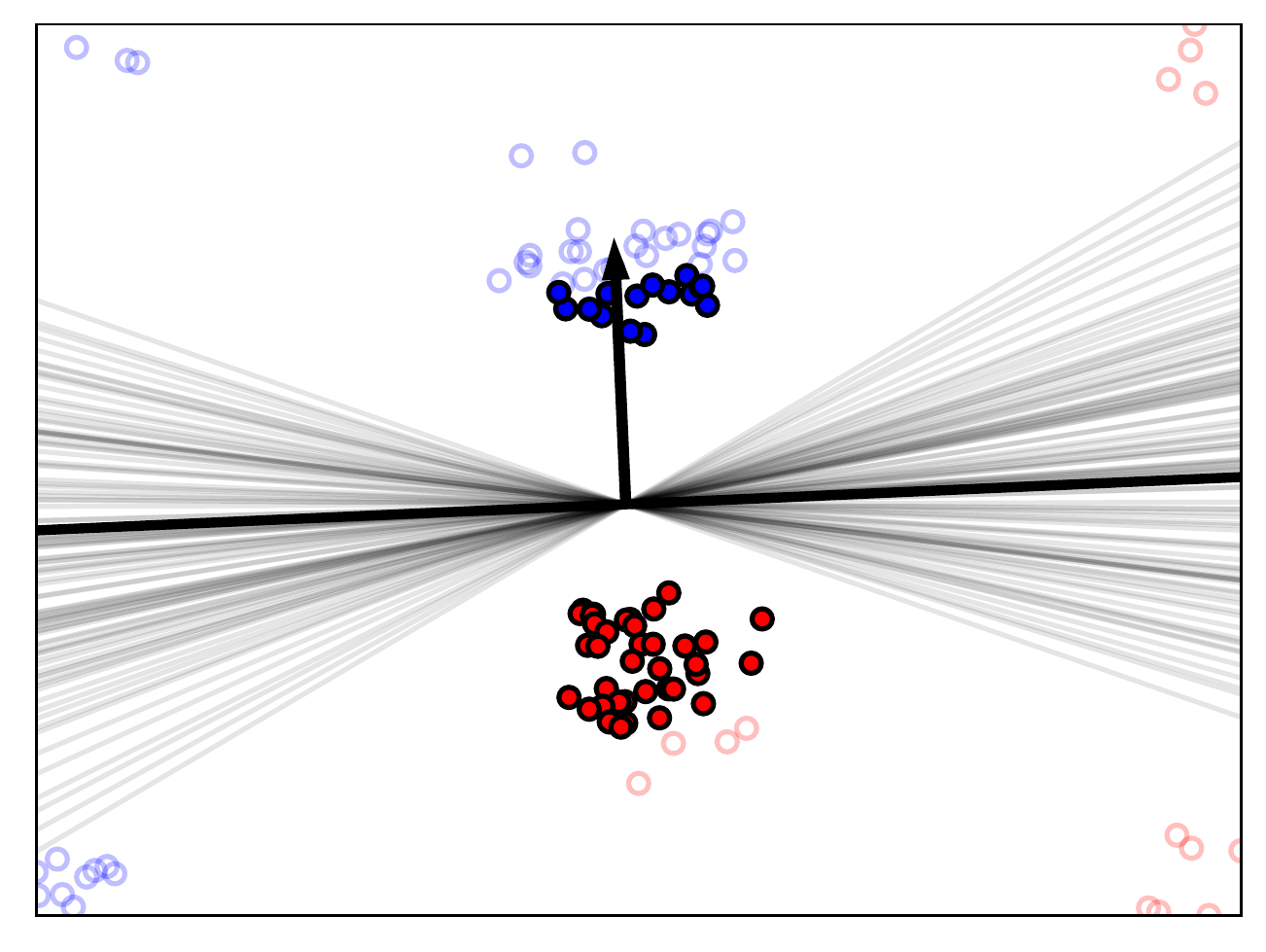}}{\small 50 Uncertainty queries}
	\end{subfigure}%
	~
	\begin{subfigure}[t]{\fw\textwidth}
		\centering
		\stackunder[2pt]{\includegraphics[width=\iw\textwidth]{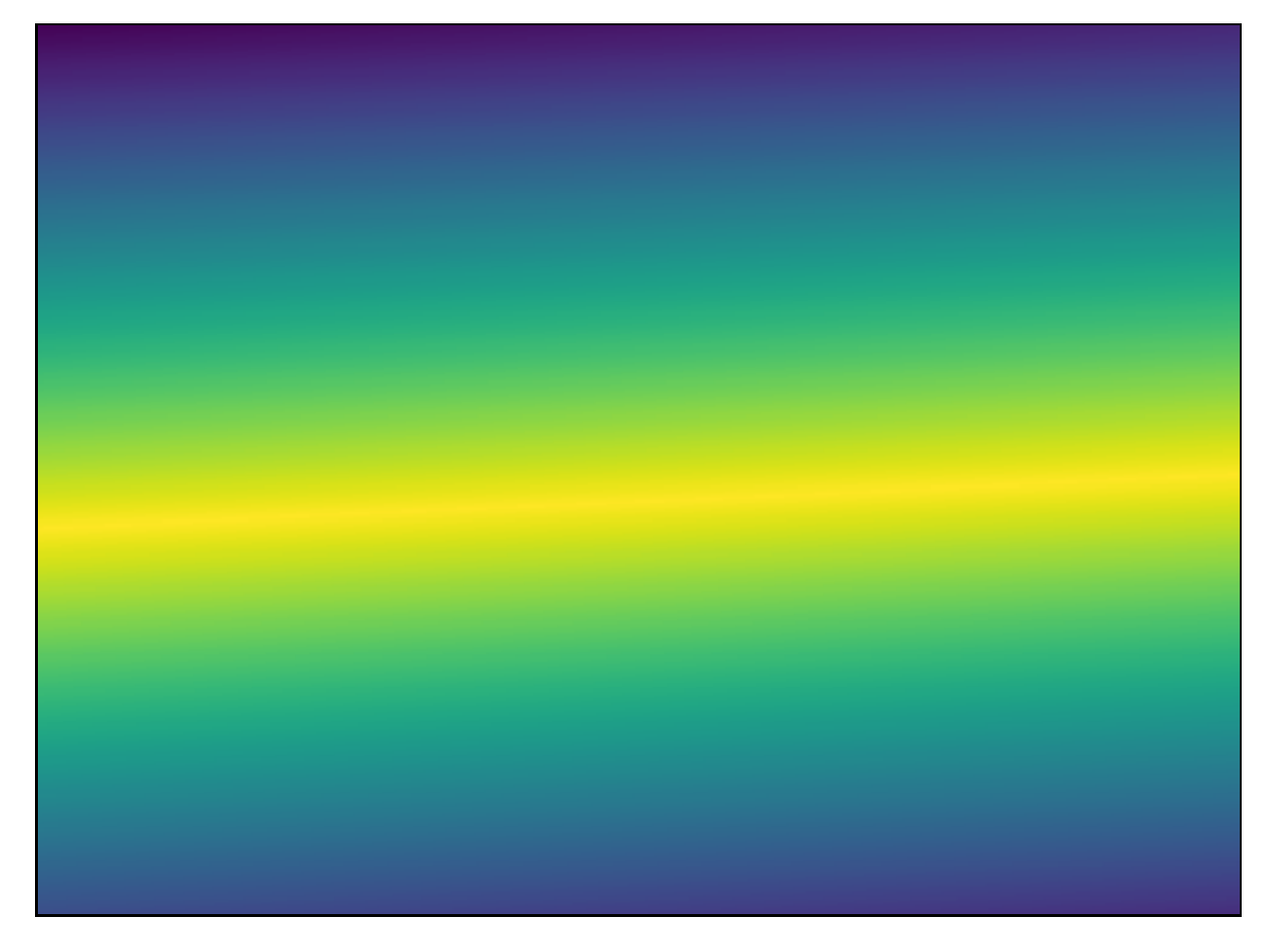}}{\small Uncertainty utility: 50 queries}
	\end{subfigure}%
	\\
	\vspace{2mm}
	\begin{subfigure}[t]{\fw\textwidth}
		\centering
		\stackunder[2pt]{\includegraphics[width=\iw\textwidth]{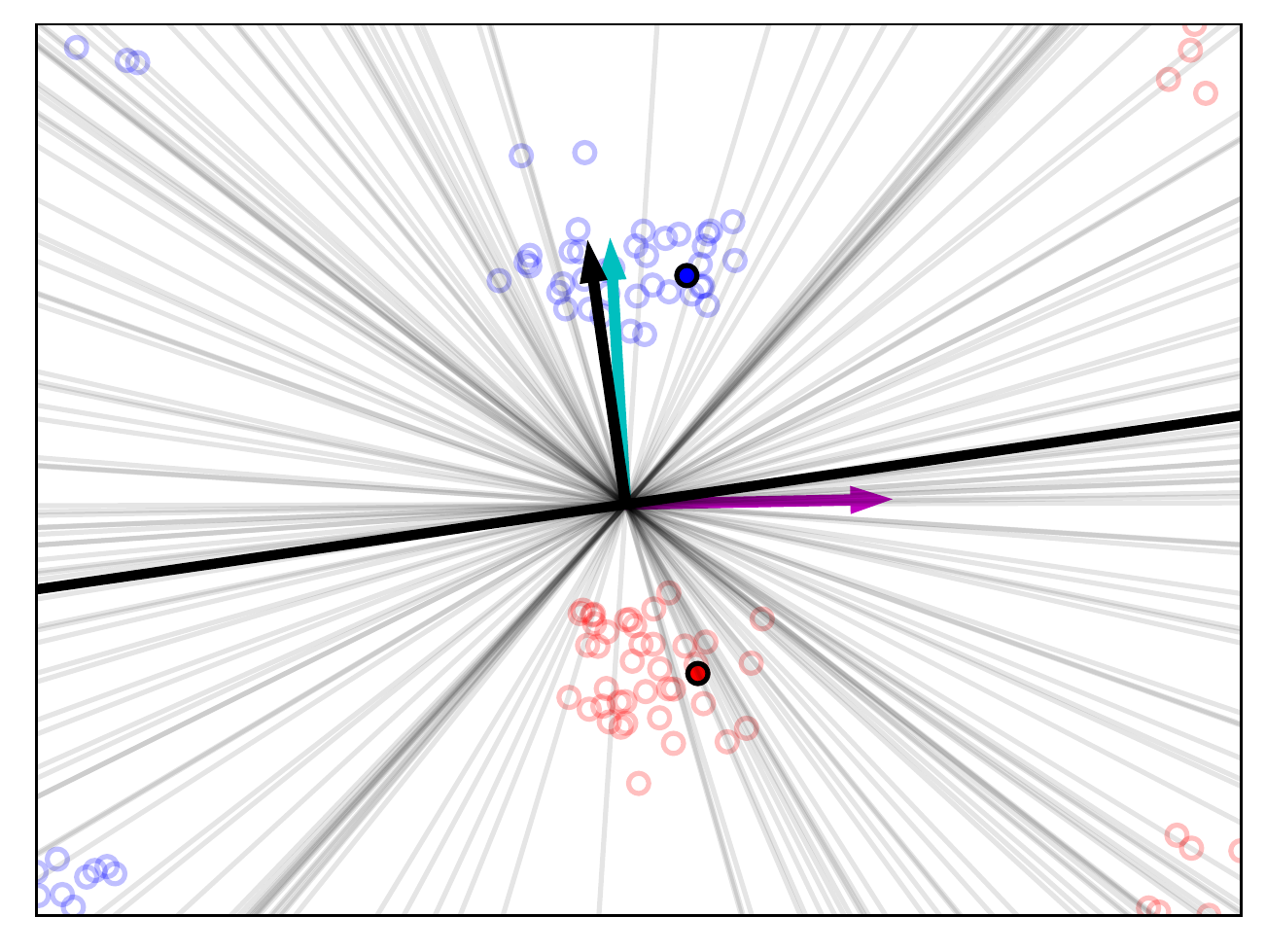}}{\small Hyperplane prior}
		\caption{}
		\label{fig:synth-a}
	\end{subfigure}%
	~
	\begin{subfigure}[t]{\fw\textwidth}
		\centering
		\stackunder[2pt]{\includegraphics[width=\iw\textwidth]{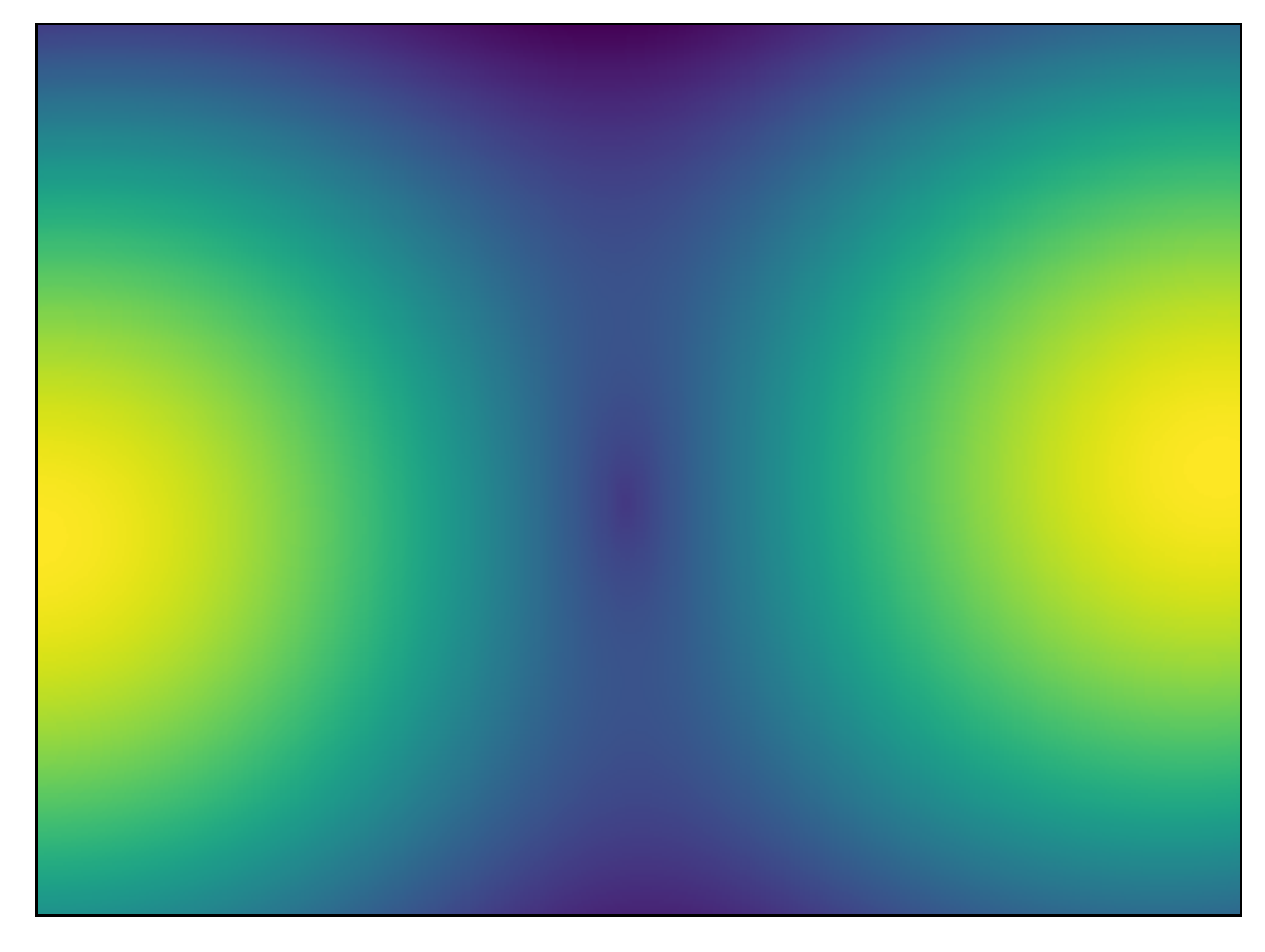}}{\small APM-LR utility: initial}
		\caption{}
		\label{fig:synth-b}
	\end{subfigure}%
	~
	\begin{subfigure}[t]{\fw\textwidth}
		\centering
		\stackunder[2pt]{\includegraphics[width=\iw\textwidth]{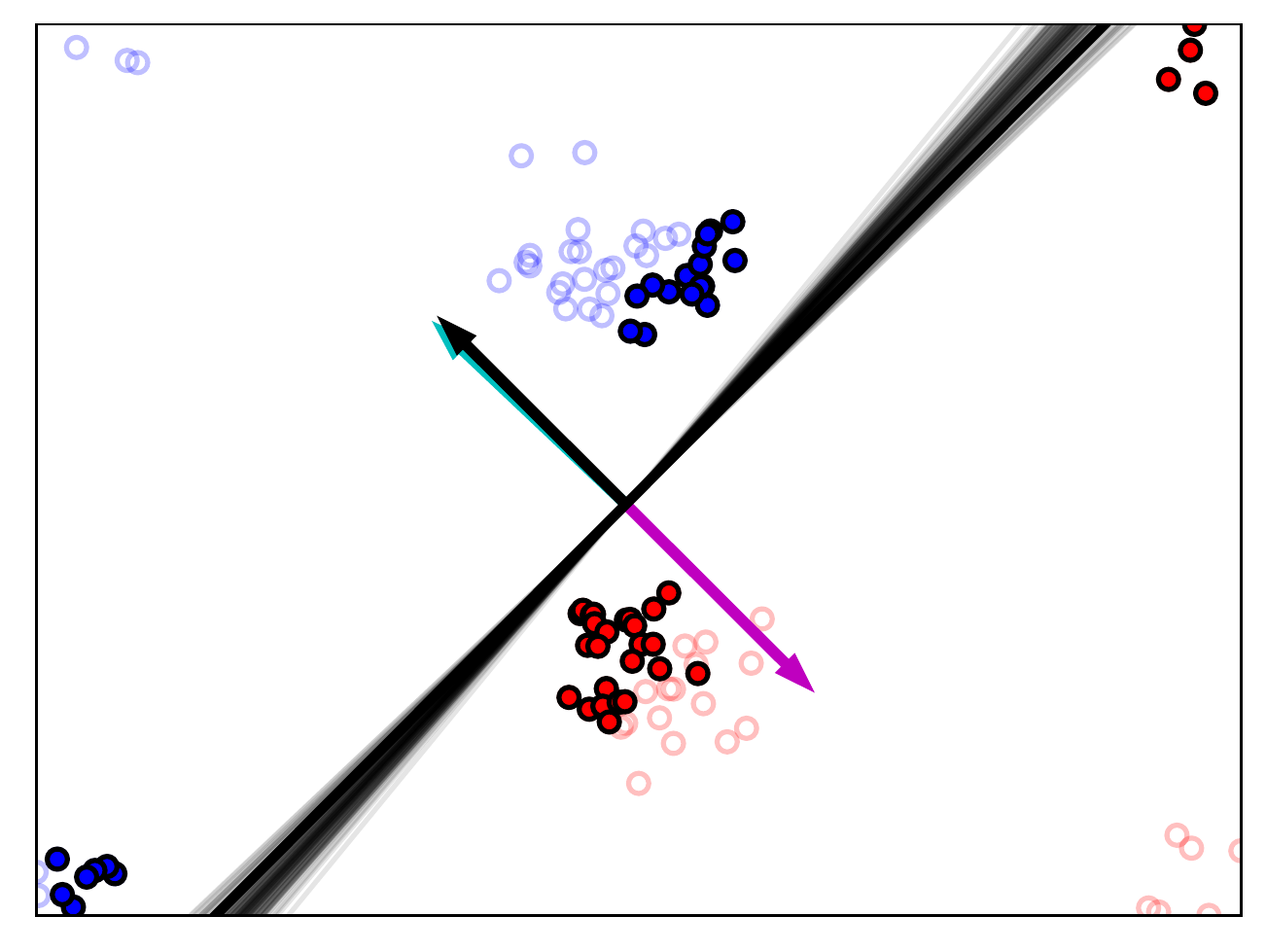}}{\small 50 APM-LR queries}
		\caption{}
		\label{fig:synth-c}
	\end{subfigure}%
	~
	\begin{subfigure}[t]{\fw\textwidth}
		\centering
		\stackunder[2pt]{\includegraphics[width=\iw\textwidth]{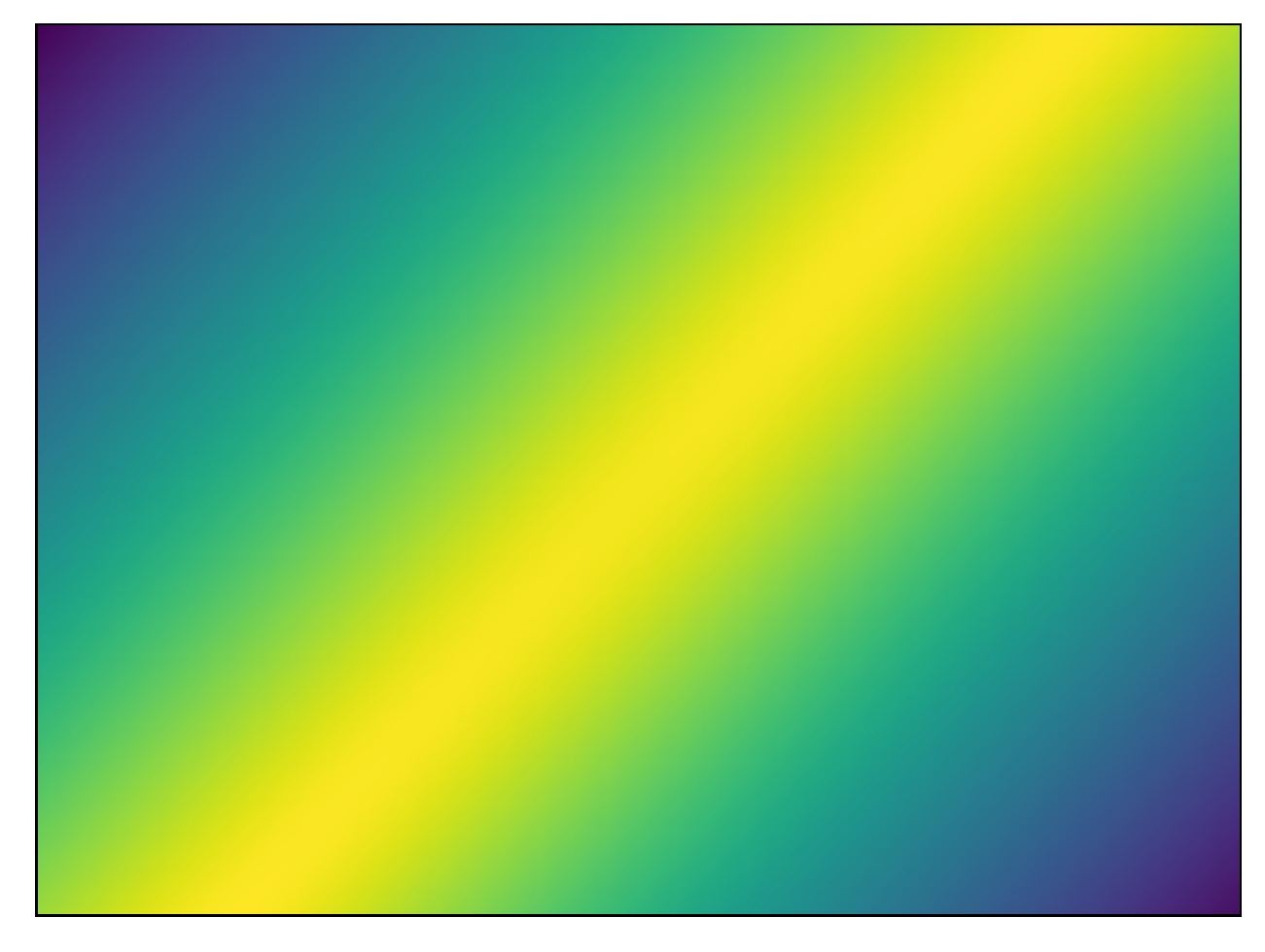}}{\small APM-LR utility: 50 queries}
		\caption{}
		\label{fig:synth-d}
	\end{subfigure}%
	\caption{(a) Top: linearly separable dataset (optimal hyperplane is diagonal) demonstrating the failure of uncertainty sampling (dataset adapted from \citet{huang2010}). Bottom: samples from hyperplane posterior, given two seed labels. The black hyperplane and corresponding normal vector depict the initial logistic regression solution, the cyan arrow indicates the normal vector to the posterior mean hyperplane, and the purple arrow indicates the maximal eigenvector of the posterior. (b) Utility function heatmap for uncertainty sampling (top) and APM-LR (bottom) --- the unlabeled example with the highest utility is selected for labeling. Uncertainty sampling selects examples close to the current hyperplane, while APM-LR selects examples that are both close to the posterior mean hyperplane and align with the direction of largest posterior variance. (c-d) After 50 queries, uncertainty sampling (c-top) has not selected samples in the dataset corners, leading to sampling bias and continued sampling of the center clusters (d-top). Meanwhile, APM-LR (c-bottom) has sufficiently explored the dataset, while continuing to sample examples in only the most ambiguous regions (d-bottom).}
	\label{fig:synth}
\end{figure*}

\subsection{Closed-form Results}

For logistic regression, the calculation of $W_2(p_L,B_t)$ takes a convenient closed-form expression, which simplifies the example selection in \eqref{eq:APM}:
\begin{prop}
	\label{prop:W2-BPSK} For $t > 0$, with $\med_{p_L}(L)$ denoting the median of $L$ according to distribution $p_L$,
	\[W_2^2(p_L,B_t) = \E_{p_L}[L^2] - 2t\E_{p_L}[\abs{L - \med_{p_L}(L)}] + t^2\]
\end{prop}
We can simplify this expression even further when $p_L$ is normally distributed.
\begin{corollary}
	\label{cor:W2-norm}
	For $L \sim \mathcal{N}(\mu,\sigma^2)$, \[W_2^2(p_L,B_t)=\mu^2 + \biggl(\sigma-\sqrt{\frac{2}{\pi}}t\biggr)^2 + \biggl(1-\frac{2}{\pi}\biggr)t^2.\]
\end{corollary}
At iteration $n$, suppose that $p_{\theta \mid \Lcal_{n-1}}$ is approximated by $\mathcal{N}(\mu_n,\Sigma_n)$, resulting in channel input $L_n = \theta^T x_n$ being distributed as $\mathcal{N}(\mu_n^T x_n, x_n^T \Sigma_n x_n)$. Although $p_{\theta \mid \Lcal_{n-1}}$ is not normally distributed in logistic regression, it is common to make this approximation in practice \citep{bishop2006}. By applying Corollary \ref{cor:W2-norm} and omitting constant terms, we derive our APM selection policy for logistic regression with power constraint $P$.
\begin{definition}
	Approximate Posterior Matching for Logistic Regression (APM-LR):
	\small
	\begin{equation}
	\pi_n(\Lcal_{n-1}) = \argmin_{x \in \Ucal_n}\enspace(\mu_n^T x)^2 + \biggl(\sqrt{x^T \Sigma_n x}-\sqrt{\frac{2}{\pi}P}\biggr)^2.\label{eq:APMLR}
	\end{equation}
\end{definition}
\normalsize
This objective is a combination of two terms: the first term corresponds to minimizing the distance between example $x$ and the posterior mean hyperplane. If $\mu_n$ is taken as an estimate of $\theta$, this term corresponds to the well-known \emph{uncertainty sampling} active learning method, which samples points close to the current hyperplane estimate \citep{tong2001support}. The second term prefers examples that align with the direction of maximum posterior covariance. Specifically, for $x^T \Sigma_n x < \frac{2}{\pi}P$, the second term is a decreasing function of $x^T \Sigma_n x$, encouraging $x$ to align with posterior covariance eigenvectors with large eigenvalues.

These two terms together can be interpreted as encouraging ``exploitation'' and ``exploration,'' respectively: the first term encourages the selection of examples that are close to the current estimate of $\theta$, exploiting this estimate to only query examples whose labels are ambiguous. The second term balances this exploitation by probing in directions of the hyperplane posterior that have not yet been sufficiently explored, reducing uncertainty about the hyperplane itself. Figure \ref{fig:synth} visualizes this tradeoff in comparison to uncertainty sampling, which only queries examples close to the current hyperplane estimate and does not account for the fact that there may be directions of the hyperplane posterior that have not been sufficiently explored. This myopic behavior is an instance of \emph{sampling bias}, a well-known phenomenon in active learning where a policy continually selects examples that reinforce the learner's belief in an incorrect hypothesis \citep{dasgupta2011,beygelzimer2009,farquhar2021on}. The balance of exploitation and exploration terms in APM-LR helps prevent this type of sampling bias, in a spirit similar to other active learning methods that balance uncertainty reduction with diverse example selection \citep{dasgupta2008,huang2010}.

An attractive computational feature of \eqref{eq:APMLR} is that the posterior mean and covariance can be estimated \emph{once} at each selection iteration and then simply projected onto each candidate example, resulting in a computational cost of only $O(d^2)$ per example evaluation. Note that these computational advantages along with the natural balance between exploration and exploitation in APM-LR emerged naturally from first-principles of feedback coding, demonstrating the potential of identifying the capacity-achieving distribution and applying APM as a universal means of designing geometrically intuitive, computationally efficient active selection schemes.
\section{Experimental Results}\label{sec:exp}
We evaluate the performance of APM-LR against baseline example selection methods for logistic regression on a variety of datasets from different tasks, as measured by holdout test accuracy and selection compute time.\footnote{Code at \url{https://github.com/siplab-gt/APM-LR}} For each method, we follow \citet{yang2018} and set the regularization parameter in \eqref{eq:LR-MAP} to $\lambda=0.01$, which we solve with the LIBLINEAR solver \citep{fan2008}. After each example is labeled, we approximate $p_{\theta \mid \Lcal_{n-1}}$ with a normal distribution by applying the variational approximation described in \citet{Jaakkola2000}, which is solved in only a few iterations of an expectation-maximization procedure (referred to here as ``VariationalEM''). The final component needed to apply APM-LR is the selection of power constraint $P$ in \eqref{eq:APMLR}.
\paragraph{Selecting Power Constraint}
Although our approach is rooted in feedback coding theory, regarding the power constraint there are two key differences between our model and traditional communications systems. First, unlike telecommunications systems that have physical restrictions such as limited battery levels, in our framework there is no external prescription of the power budget $P$ and therefore we can select any valid upper bound on the channel input power induced by the unlabeled examples. Secondly, unlike coding schemes which globally maximize information gain over the entire trajectory of channel inputs, we seek to myopically maximize the one-step information gain at every channel input. Since the goal at each iteration is to separately solve a local information maximization problem, there is no need for the power constraint $P$ to be constant across iterations, and therefore we set a separate power constraint $P_n$ for each iteration.

Since the selection of $P_n$ parameterizes the target distribution in APM-LR, it is important for $P_n$ to be set as tight as possible so that the target capacity-achieving distribution is well-matched to the set of feasible channel input distributions. This is because at each iteration the capacity-achieving distribution serves as a proxy for the optimal input distribution induced by a real example, and a setting of $P_n$ that is too loose will result in APM targeting a proxy that is not well-matched to the feasible input distributions. To select a satisfactory setting of $P_n$, we derive an upper bound on the channel input power to use as an implicit constraint.

\begin{wrapfigure}{R}{0.5\textwidth}
	\begin{minipage}{0.5\textwidth}
		\begin{algorithm}[H]
			\caption{Approximate Posterior Matching for Logistic Regression (APM-LR)}
			\label{alg:APMLR}
			\begin{algorithmic}[1]
				\REQUIRE data pool $\X$, hyperparameter $\lambda > 0$, horizon $N$, initial training set $\Lcal$
				\STATE $\mu \gets 0$, $\Sigma \gets \frac1\lambda I$
				\STATE $B \gets \max_{x \in \Ucal} \norm{x}_2$
				\STATE $\Ucal \gets \X$
				\FOR{$n = 1$ \TO $N$}
				\STATE $P \gets B^2 \lambda_1(\Sigma)$
				\STATE $x^* \gets \argmin_{x \in \Ucal}(\mu^T x)^2 + \Bigl(\sqrt{x^T \Sigma x}-\sqrt{\frac{2}{\pi}P} \, \Bigr)^2$
				\STATE $y^* \gets \operatorname{ExpertLabel}(x^*)$
				\STATE $\Ucal \gets \Ucal \setminus \{x^*\}$, $\Lcal \gets \Lcal \cup (x^*,y^*)$
				\STATE $\mu,\Sigma \gets \operatorname{VariationalEM}(\Lcal)$
				\STATE $\theta^* \gets \A(\Lcal)$ i.e., eq.\ \eqref{eq:LR-MAP}
				\ENDFOR
				\ENSURE hyperplane $\theta^*$
			\end{algorithmic}
		\end{algorithm}
	\end{minipage}
\end{wrapfigure}

Suppose for a given dataset that there exists a known $B > 0$ such that $\norm{x}_2 < B$ (this is a reasonable assumption in many real-world settings). Let $\lambda_1(M)$ denote the largest magnitude eigenvalue of matrix $M$. We then have (with expectations taken with respect to $p_{L_n \mid \Lcal_{n-1}}$)
\[
\E[L_n^2] = x^T (\mu_n \mu_n^T + \Sigma_n) x \le B^2 \lambda_1(\mu_n \mu_n^T + \Sigma_n).
\]
For each $n$ we can therefore set $P_n = B^2 \lambda_1(\mu_n \mu_n^T + \Sigma_n)$. In our experiments we select a slightly modified parameter $P_n=B^2 \lambda_1(\Sigma_n)$, which we justify as a more practical heuristic in Appendix \ref{app:power}. We summarize APM-LR in full in Algorithm \ref{alg:APMLR}, including power constraint calculation and variational posterior updating.

\paragraph{Datasets}
We follow previous work in active learning for logistic regression \citep{huang2010,yang2018} and test each method on several UCI datasets \citep{dua2017} including \emph{vehicle}, \emph{letter}, \emph{austra}, and \emph{wdbc}. We also evaluate performance on several synthetic datasets including the dataset depicted in Figure \ref{fig:synth} (adapted from \citet{huang2010}), which we refer to as \emph{cross} (see Appendix \ref{app:data} for details on all datasets). For each simulation trial, we first randomly divide the dataset into an equally-sized data pool ($\Ucal$) and held-out test set. We normalize $\Ucal$ to zero-mean and coordinate-wise unit-variance, and apply the same transformation to the test set. Before evaluating each example selection method, the training dataset ($\Lcal$) is seeded to consist of one randomly selected labeled example from each class.\footnote{Our experiments are synchronized across data selection methods: each trial uses the same training/test split and seed examples for each tested method.}

\paragraph{Baseline Methods} 
We evaluate the following baseline methods, each described with their computational cost per candidate example evaluation (see Appendix \ref{app:methods} for details):
\begin{itemize}[leftmargin=*,topsep=0pt,itemsep=5pt,parsep=0pt]
	\item \emph{Uncertainty}: select closest example to current hyperplane estimate (i.e.\ $\argmin_{x \in \Ucal_n} x^T \widehat{\theta}_{n-1}$) at cost $O(d)$. The action of Uncertainty sampling is comparable to that of the first term in \eqref{eq:APMLR}.
	\item \emph{Random}: each example is selected uniformly at random from $\Ucal_n$, at $O(1)$ cost.
	\item \emph{MaxVar}: to isolate the effect of the second term in \eqref{eq:APMLR}, we evaluate a control strategy that selects the example that induces the largest channel input variance (i.e.\ $\argmax_{x \in \Ucal_n} x^T \Sigma_n x$), at cost $O(d^2)$.
	\item \emph{InfoGain}: selects the example with the largest information gain $I(\theta; Y_n \mid \Lcal_{n-1})$, estimated by sampling $s$ times from the normally approximated hyperplane posterior (here we set $s=100$) and for each candidate example evaluating a Monte Carlo approximation of information gain, at $O(d s)$ cost.
	\item \emph{BALD}: we approximate the logistic function $f(\ell)$ with a probit function and apply the probit regression active learning method of \citet{houlsby2011bayesian}, at cost $O(d^2)$. Like APM-LR, BALD approximates the action of InfoGain and only requires the mean and covariance of the normally approximated hyperplane posterior.
\end{itemize}
InfoGain is the most computationally intensive selection method, since it requires a brute-force Monte Carlo approximation of information gain for each candidate example. BALD and APM-LR have the next least expensive cost per example at $O(d^2)$, followed by Uncertainty and Random sampling.
\begin{figure*}[t]
	\centering
	\begin{subfigure}[t]{0.33\textwidth}
		\centering
		\includegraphics[width=\textwidth]{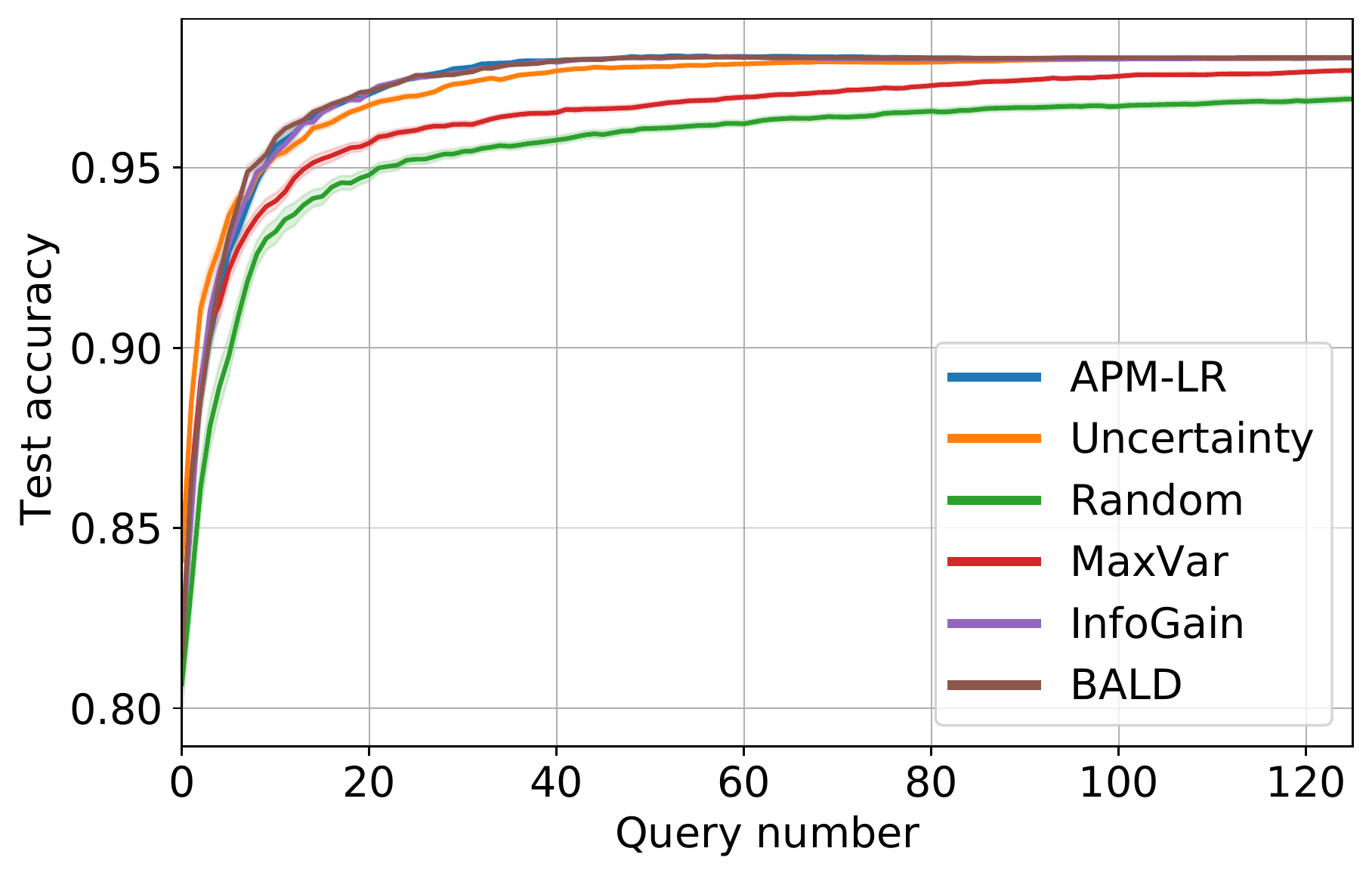}
		\caption{\emph{letterDP}}
		\label{fig:exper-letter-dvp}
	\end{subfigure}%
	~
	\begin{subfigure}[t]{0.33\textwidth}
		\centering
		\includegraphics[width=\textwidth]{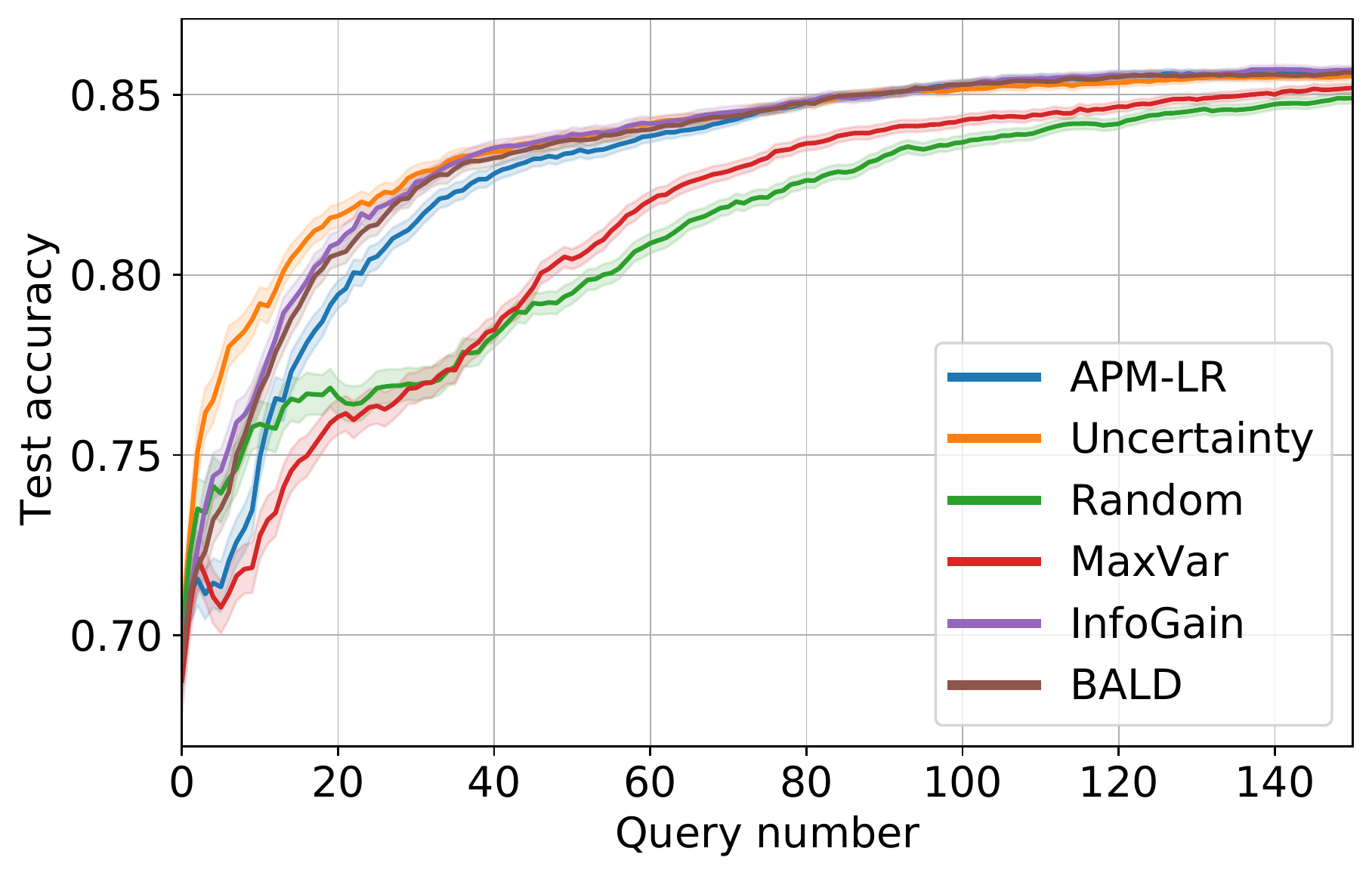}
		\caption{\emph{austra}}
		\label{fig:exper-australian}
	\end{subfigure}%
	~
	\begin{subfigure}[t]{0.33\textwidth}
		\centering
		\includegraphics[width=\textwidth]{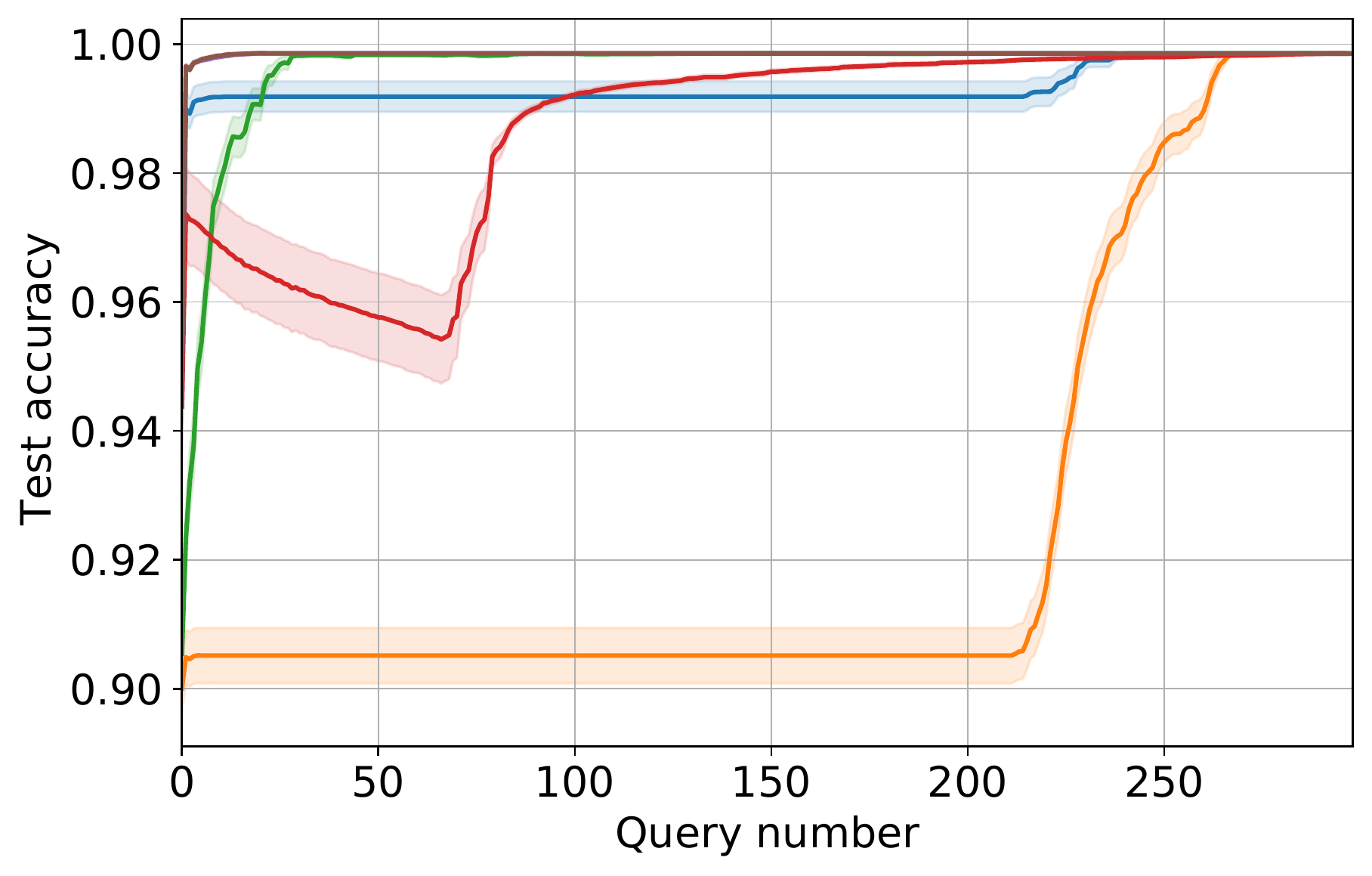}
		\caption{\emph{cross}}
		\label{fig:exper-cross}
	\end{subfigure}%
	\caption{Average test classification accuracy plotted against number of labeled examples (error bars show $\pm 1$ standard error) across select UCI datasets (a-b) and the synthetic \emph{cross} dataset (c, with legend shared with a-b and omitted for visual clarity). Overall, APM-LR performs comparably to other methods seeking to approximately maximize information gain. While uncertainty sampling performs well on some datasets (a-b), it can fail in cases where it suffers from sampling bias (c). Most of the tested active learning methods (except the control, MaxVar) outperform random sampling. For visual clarity we show different numbers of queried examples for each dataset.}
	\label{fig:exper}
\end{figure*}
\paragraph{Performance Comparison}
In Figure \ref{fig:exper}, we compare the learning performance of each data selection method  by plotting holdout test accuracy against number of queried examples (excluding the seed set) across select datasets (see Appendix \ref{app:test} for full results). We generally find that the tested active data selection methods outperform random sampling. The exception is MaxVar, which performs comparably to random selection and worse than APM-LR. Although simple Uncertainty sampling matches the performance of other active methods on several datasets (Figure \ref{fig:exper}a-b) as previously observed by \citeauthor{yang2018}, in additional tests on synthetic datasets we find that APM-LR outperforms uncertainty sampling. This is the case for the \emph{cross} dataset (Figure \ref{fig:exper-cross}), demonstrating how Uncertainty sampling can be susceptible to sampling bias that leads to insufficient exploration (see Appendix \ref{app:fail} for additional failure mode analysis). These tests together lend evidence to the mixture of terms in \eqref{eq:APMLR} having combined benefits over pure exploration of directions with large posterior variance or pure exploitation of ambiguous examples with respect to the current hyperplane estimate. Finally, APM-LR generally performs similarly to InfoGain and BALD, both of which directly approximate the action of information gain maximization, in contrast to APM's geometric, indirect approach.

\begin{wraptable}{R}{0.5\textwidth}
	\centering
	\begin{small}
\begin{tabular}{|>{\centering}p{15mm}|S[table-format=2.3]|S[table-format=2.3]|S[table-format=2.3]|}
\hline
{} & {\emph{letterDP}} & {\emph{austra}} & {\emph{cross}} \\  
\hline
{APM-LR} & 0.336 & 0.150 & 0.125 \\  
\hline
{Uncertainty} & 0.149 & 0.063 & 0.053 \\  
\hline
{BALD} & 4.230 & 1.770 & 1.521 \\  
\hline
{InfoGain} & 12.755 & 5.089 & 2.722 \\  
\hline
{Random} & 0.005 & 0.003 & 0.002 \\  
\hline
{MaxVar} & 0.118 & 0.050 & 0.040 \\  
\hline
\end{tabular}
\end{small}

	\caption{Comparison of median cumulative time (s) for each method to select the first 40 examples (excluding seed points and time for model retraining). Generally, APM-LR has a cost an order of magnitude lower than InfoGain and BALD (which  directly approximate the action of information maximization), while Uncertainty, MaxVar, and Random sampling have the cheapest cost.}
	\label{tab:compute-cost}
\end{wraptable}
Table \ref{tab:compute-cost} depicts the computational cost for each method across select datasets (see Appendix \ref{app:comp} for full results and expanded timing evaluations). Similar to the analysis in \citet{yang2018}, for each method we evaluate the cumulative compute time to select the first 40 examples (excluding seed examples and time for model retraining), and compute the median time over all trials. We see that InfoGain is the most expensive of all methods, since it directly approximates information gain with Monte Carlo sampling. BALD has the next highest cost, followed by APM-LR --- the two latter methods only require a single computation of posterior mean and variance, which can be projected onto each candidate example. Uncertainty sampling and random sampling have the lowest computational cost.

Although BALD can also be computed using only the posterior mean and covariance, it is unclear how the approximation in BALD can be applied beyond probit regression. In contrast, the APM formulation in \eqref{eq:APM} can be applied generally to \emph{any} active learning problem that can be decomposed into a deterministic encoder and noisy channel, along with a known capacity-achieving distribution. The combined results of Figure \ref{fig:exper} and Table \ref{tab:compute-cost} suggest that the universal APM approach of leveraging this analytical knowledge of the capacity-achieving distribution affords a geometric active selection approach that performs well in terms of both sample and computational complexity.

\section{Conclusion}
To our knowledge, our work is the first effort to both reframe active learning as a feedback communications system and utilize analytical knowledge of the corresponding capacity-achieving distribution to derive an active learning scheme. The analytical and empirical results in this work for the special case of logistic regression demonstrate the potential of this coding-based active learning approach: information continuity results show how examples selected with APM-LR have information gain approaching their maximum possible value, APM-LR has a convenient geometrical formulation resulting from analytical knowledge of the capacity-achieving distribution for logistic regression (characterized here for the first time) that can lead to computationally efficient example selection, and when tested on multiple datasets APM-LR performs comparably to baseline active learning methods including brute-force information maximization. APM-LR's attractive balance between exploration and exploitation emerged naturally from first-principles of channel coding, extending beyond the common approach of uncertainty sampling. 

More generally, a fundamental feature of Approximate Posterior Matching is that analytical knowledge of the capacity-achieving distribution converts the usually unwieldy \emph{information maximization} problem in active learning to a \emph{geometric} problem. In logistic regression, this geometry led to computational advantages over direct information maximization, and we conjecture that similar benefits may emerge in more complex settings. Additionally, the general formulation of APM in \eqref{eq:APM} presents several opportunities to leverage existing computational algorithms to aid example selection, including estimating $p^*_L$ when it is analytically unknown \citep{blahut1972, arimoto1972} and optimizing Wasserstein distances with state-of-the-art methods \citep{peyre2019}. Overall, we believe that our coding-theoretic approach opens several new directions for future work in active learning.

\section*{Acknowledgements}
We thank the reviewers for their useful feedback and comments, as well as Rob Nowak, John Lee, and other colleagues for insightful discussions. This work was supported by NSF grant CCF-1350954, ONR grant N00014-15-1-2619, and the IDEaS-TRIAD Research Scholarship under NSF grant 1740776. This research was also supported in part through research cyberinfrastructure resources and services provided by the Partnership for an Advanced Computing Environment (PACE) at the Georgia Institute of Technology, Atlanta, Georgia, USA.

\bibliography{CodingALarxiv}

\appendix\clearpage
\section{Proofs of Analytical Results}
\label{app:proofs}

\subsection{Proof of Proposition 2.1}
\begin{proof}
	Our proof follows closely to that of \citet{singh2009} for the capacity of the one-bit quantized Gaussian channel. We start by writing $I(L;Y) = H(Y) - H(Y \mid L)$, where $H$ denotes the entropy of a discrete random variable \citep{cover2006}. $H(Y)$ is maximized at 1 bit, when $p(Y = 1) = p(Y = -1) = 0.5$. Expanding $H(Y \mid L)$, we have $H(Y \mid L) = \E_{p_L}[h_b(p(Y = 1 \mid L))] = \E_{p_L}[h_b(f(L))]$.
	
	For distribution $p_L$, consider its symmetrized distribution $\widetilde{p}_L(\ell) = \frac12 p_L(\ell) + \frac12 p_L(-\ell)$ and the expectation of any even function $e(\cdot)$ over $\widetilde{p}_L(\ell)$:
	\begin{align*}
	E_{\widetilde{p}_L}[e(L)] &= \int_{-\infty}^\infty \biggl(\frac12 p_L(\ell) + \frac12 p_L(-\ell)\biggr) e(\ell) d\ell
	\\ &= \frac12 \int_{-\infty}^\infty p_L(\ell) e(\ell) d\ell + \frac12 \int_{-\infty}^\infty p_L(-\ell) e(\ell) d\ell
	\\ &= \frac12 \int_{-\infty}^\infty p_L(\ell) e(\ell) d\ell + \frac12 \int_{-\infty}^\infty p_L(\ell) e(-\ell) d\ell &\text{change of variables}
	\\ &= \frac12 \int_{-\infty}^\infty p_L(\ell) e(\ell) d\ell + \frac12 \int_{-\infty}^\infty p_L(\ell) e(\ell) d\ell &\text{ $e(\ell)$ is even}
	\\ &= \frac12 E_{p_L}[e(L)] + \frac12 E_{p_L}[e(L)]
	\\ &= E_{p_L}[e(L)]
	\end{align*}	
	Observe that $h_b$ is symmetric about $0.5$, i.e.\ for $x \in [-0.5,0.5]$, $h_b(0.5 + x) = h_b(0.5-x)$. Combining this with the fact that $f(\ell) - 0.5$ is an odd function (i.e.\ $f(-\ell) - 0.5 = -(f(\ell) - 0.5))$, we have
	\[h_b(f(-\ell)) = h_b(f(-\ell) - 0.5 + 0.5) =  h_b(-(f(\ell) - 0.5) + 0.5) = h_b((f(\ell) - 0.5) + 0.5) = h_b(f(\ell))\]
	and so $h_b(f(\ell))$ is an even function. Therefore, the conditional entropy $H(Y \mid L)$ is equivalent when $L$ is distributed as $p_L$ or $\widetilde{p}_L$, i.e.\ $E_{\widetilde{p}_L}[h_b(f(L))]=E_{p_L}[h_b(f(L))]$.
	
	We also have
	\begin{align*}
	E_{\widetilde{p}_L}[f(L)] &= E_{\widetilde{p}_L}[f(L) - 0.5] + 0.5
	\\ &= \int_{-\infty}^\infty \biggl(\frac12 p_L(\ell) + \frac12 p_L(-\ell)\biggr) (f(\ell)-0.5) d\ell + 0.5
	\\ &= \frac12 \int_{-\infty}^\infty p_L(\ell) (f(\ell)-0.5) d\ell + \frac12 \int_{-\infty}^\infty p_L(-\ell) (f(\ell)-0.5) d\ell + 0.5
	\\ &= \frac12 \int_{-\infty}^\infty p_L(\ell) (f(\ell)-0.5) d\ell + \frac12 \int_{-\infty}^\infty p_L(\ell) (f(-\ell)-0.5) d\ell +0.5 &\text{change of variables}
	\\ &= \frac12 \int_{-\infty}^\infty p_L(\ell) (f(\ell)-0.5) d\ell - \frac12 \int_{-\infty}^\infty p_L(\ell) (f(\ell)-0.5) d\ell +0.5&\text{ $(f(\ell)-0.5)$ is odd}
	\\ &= 0.5
	\end{align*}
	and so under $\widetilde{p}_L$, $p(Y=1) = \E_{\widetilde{p}_L}[f(L)] = 0.5$ and $H(Y)$ is maximized at 1 bit.
	
	Combining these facts, we have
	\[I(\widetilde{p}_L, f) = 1 - E_{\widetilde{p}_L}[h_b(f(L))] = 1 - E_{p_L}[h_b(f(L))] \ge h_b(E_{p_L}[f(L)]) - E_{p_L}[h_b(f(L))] = I(p_L, f)\]
	and so symmetrizing a distribution can only increase $I(L;Y)$. Furthermore, since $\ell^2$ is even we have $\E_{\widetilde{p}_L}[L^2] = \E_{p_L}[L^2]$. Therefore, when evaluating the capacity of channel with transition probability $f$ under power constraint $P$, we only consider symmetric distributions since for every $p_L \in \C_P$ there exists a symmetric distribution $\widetilde{p}_L \in \C_P$ satisfying $I(\widetilde{p}_L,f) \ge I(p_L, f)$. We solve for the capacity-achieving distribution over the set of symmetric distributions in $\C_P$:
	\begin{align}
	p_L^* &= \argmax_{\substack{\E_{p_L}[L^2] \le P \\ p_L(\ell) = p_L(-\ell)}} I(p_L, f)\label{eq:max-info}
	\\ &= \argmax_{\substack{\E_{p_L}[L^2] \le P \\ p_L(\ell) = p_L(-\ell)}} 1 - E_{p_L}[h_b(f(L))]\notag
	\\ &= \argmin_{\substack{\E_{p_L}[L^2] \le P \\ p_L(\ell) = p_L(-\ell)}} E_{p_L}[h_b(f(L))]\label{eq:min-cond-ent}
	\end{align}
	
	Since $h_b(f(\ell))$ is even, $h_b(f(\ell)) = h_b(f(\abs{\ell})) = h_b(f(\sqrt{\ell^2}))$. Omitting calculations, we have
	\[\frac{d^2}{du^2} h_b(f(\sqrt{u})) = (\log_2 e)\frac{\tanh(\frac{\sqrt{u}}{2})\operatorname{sech}^2(\frac{\sqrt{u}}{2})}{16 \sqrt{u}}\]
	which is non-negative for $u > 0$ and therefore $h_b(f(\sqrt{u}))$ (which is continuous on $u \ge 0)$ is convex on $u \ge 0$. We then have
	\[E_{p_L}[h_b(f(L))]=E_{p_L}\bigl[h_b\bigl(f\bigl(\sqrt{L^2}\bigr)\bigr)\bigr] \stackrel{(a)}\ge h_b\Bigl(f\Bigl(\sqrt{E_{p_L}[L^2]}\Bigr)\Bigr) \stackrel{(b)}\ge h_b\bigl(f\bigl(\sqrt{P}\bigr)\bigr)\]
	where Jensen's inequality is used in (a) \citep{cover2006}, with equality if and only if $L^2$ is constant, and (b) results from the power constraint $E_{p_L}[L^2] \le P$ and the fact that $h_b(f(\sqrt{u}))$ is monotonically decreasing for $u \ge 0$. For symmetric $p_L$, equality in (a) is achieved if $p_L = B_t$ for some $t > 0$. By setting $t = \sqrt{P}$, equality in (b) is also achieved, and so $B_{\sqrt{P}}$ minimizes \eqref{eq:min-cond-ent} (and therefore maximizes \eqref{eq:max-info}). The maximum value in \eqref{eq:max-info}, which is equal to capacity $C$, is then \[I(B_{\sqrt{P}},f) = 1 - \E_{B_{\sqrt{P}}}[h_b(f(L))] = 1 - \frac12 h_b(f(\sqrt{P})) - \frac12 h_b(f(-\sqrt{P})) = 1 - h_b(f(\sqrt{P})).\qedhere\]
\end{proof}

\subsection{Proof of Proposition 2.2}
\begin{proof}
	Since $p_\theta$ is log-concave, then $p_{\theta \mid \Lcal_{n-1}}(\theta) \propto p_\theta(\theta) \prod_{i=1}^{n-1} p(Y = y_i \mid x_i, \theta)$ is also log-concave since it is the product of log-concave functions \citep{saumard2014}. Since marginals of log-concave distributions are log-concave \citep{lovasz2007}, $L_n = x_n^T \theta$ is log-concave for any $x_n$ under the distribution $p_{\theta \mid \Lcal{n-1}}$. However, we know from Proposition 2.1 that $p^*_L$ for logistic regression is a sum of mass points, which is not log-concave. Therefore no $x_n$ exists which can induce $p^*_L$ from $h$.
\end{proof}

\subsection{Proof of Theorem 3.1}
\begin{proof}
	In the following, suppose that $p_L \in \C_P$, and let $H_{p_L}(Y) = h_b(\E_{p_L}[f(L)])$ and $H_{p_L}(Y \mid L) = \E_{p_L}[h_b(f(L))]$. $f(\ell)$ is $K_1$-Lipschitz, where $K_1 = 0.25$, and $h_b(f(\ell))$ is $K_2$-Lipschitz, where $K_2 \approx 0.32$.
	\begin{align*}
	\abs{I(p_L,f) - I(B_t,f)} &= \abs{H_{p_L}(Y) - H_{p_L}(Y \mid L) - (H_{B_t}(Y) - H_{B_t}(Y \mid L))}
	\\ &\le \abs{H_{p_L}(Y) - H_{B_t}(Y))} + \abs{H_{p_L}(Y \mid L) - H_{B_t}(Y \mid L)}
	\\ &= \abs{h_b(\E_{p_L}[f(L)]) - h_b(\E_{B_t}[f(L)])} + \biggl\lvert\int_\ell h_b(f(\ell)) p_L(\ell) d\ell - \int_\ell h_b(f(\ell)) B_t(\ell) d\ell\biggr\rvert
	\end{align*}
	Assume that there exists $\varepsilon \in (0,0.5)$ such that $\varepsilon \le \E_{p_L}[f(L)] \le 1-\varepsilon$. For $\ell \in (\varepsilon,1-\varepsilon)$, $h_b$ is $\log_2\frac{1-\varepsilon}{\varepsilon}$-Lipschitz.
	Since $\varepsilon < \E_{p_L}[f(L)] < 1-\varepsilon$ by assumption and $B_t$ satisfies $\varepsilon < \E_{B_t}[f(L)] < 1-\varepsilon$ since $\E_{B_t}[f(L)] = 0.5$, we have
	\begin{align*}
	\abs{h_b(\E_{p_L}[f(L)]) - h_b(\E_{B_t}[f(L)])} &\le \log_2\Bigl(\frac{1-\varepsilon}{\varepsilon}\Bigr)\abs{\E_{p_L}[f(L)] - \E_{B_t}[f(L)]}
	\\ &=\log_2\Bigl(\frac{1-\varepsilon}{\varepsilon}\Bigr)\biggl\lvert\int_\ell f(\ell) p_L(\ell) d\ell - \int_\ell f(\ell) B_t(\ell) d\ell\biggr\rvert
	\end{align*}
	which implies
	\begin{small}
		\begin{equation}
		\abs{I(p_L,f) - I(B_t,f)}  \le \log_2\Bigl(\frac{1-\varepsilon}{\varepsilon}\Bigr)\biggl\lvert\int_\ell f(\ell) p_L(\ell) d\ell - \int_\ell f(\ell) B_t(\ell) d\ell\biggr\rvert + \biggl\lvert\int_\ell h_b(f(\ell)) p_L(\ell) d\ell - \int_\ell h_b(f(\ell)) B_t(\ell) d\ell\biggr\rvert\label{eq:info-upper-crude}
		\end{equation}
	\end{small}
	
	To continue, we use the following result from \citet{villani2008optimal}: defining $P_1(\R) \coloneqq \{\mu' : \E_{\mu'}[\abs{L}] < \infty\}$, for any $\mu,\nu \in P_1(\R)$ we have
	\[\sup_{\norm{f}_{\mathrm{Lip}}\le 1}\int_{\ell} f(\ell) \mu(\ell) d\ell - \int_{\ell} f(\ell) \nu(\ell) d\ell = W_1(\mu,\nu).\]
	Therefore, for any $K$-Lipschitz function $g$ we have that $\frac{g}{K}$ is 1-Lipschitz and so
	\begin{align}
	\biggl\lvert\int_{\ell} g(\ell) \mu(\ell) d\ell - \int_{\ell} g(\ell) \nu(\ell) d\ell \biggr\rvert&= K\biggl\lvert\int_{\ell} \frac{g(\ell)}{K} \mu(\ell) d\ell - \int_{\ell} \frac{g(\ell)}{K} \nu(\ell) d\ell\biggr\rvert \notag
	\\ &= K\max\biggl\{\int_{\ell} \frac{g(\ell)}{K} \mu(\ell) d\ell - \int_{\ell} \frac{g(\ell)}{K} \nu(\ell) d\ell, \int_{\ell} \frac{-g(\ell)}{K} \mu(\ell) d\ell - \int_{\ell} \frac{-g(\ell)}{K} \nu(\ell) d\ell\biggr\} \notag
	\\ &\le K \sup_{\norm{f}_{\mathrm{Lip}}\le 1}\int_{\ell} f(\ell) \mu(\ell) d\ell - \int_{\ell} f(\ell) \nu(\ell) d\ell \notag
	\\ &\le K W_1(\mu,\nu) \notag
	\\ &\le K W_2(\mu,\nu)\label{eq:inequal-W1W2}
	\end{align}
	where the last inequality is from $W_1(\mu,\nu) \le W_2(\mu,\nu)$ \citep{villani2008optimal}.
	
	To apply this inequality to both expressions in \eqref{eq:info-upper-crude}, we first verify that $p_L, B_t \in P_1(\R)$. $\E_{B_t}[\abs{L}] = t < \infty$, and \[\E_{p_L}[\abs{L}] = \E_{p_L}\Bigl[\sqrt{L^2}\Bigr] \stackrel{(a)}{\le} \sqrt{\E_{p_L}[L^2]} \stackrel{(b)}{\le} \sqrt{P} < \infty\]
	where (a) results from Jensen's inequality with the concavity of $\sqrt{\cdot}$, and (b) is since $\E_{p_L}[L^2] \le P$ by assumption and $\sqrt{\cdot}$ is monotonically increasing. Applying \eqref{eq:inequal-W1W2} separately to both terms in \eqref{eq:info-upper-crude}, we have
	\begin{equation}
	\abs{I(p_L,f) - I(B_t,f)}  \le \biggl(K_1 \log_2\Bigl(\frac{1-\varepsilon}{\varepsilon}\Bigr) + K_2 \biggr) W_2(p_L,B_t)\label{eq:W2upper-eps}
	\end{equation}
	
	Finally, we compute a valid value of $\varepsilon$ for all $p_L \in \mathcal{C}_P$. First note that $f(\ell) < 0.5$ for $\ell < 0$ and $f(\ell) \ge 0.5$ for $\ell \ge 0$, implying that $f(\ell) \le f(\abs{\ell})\;\forall \ell$. Next note that $f(\sqrt{u})$ is concave on $u \ge 0$, since for any $u,v \in [0,\infty)$ and any $0 < \phi < 1$
	\begin{align*}
	f(\sqrt{\phi u + (1-\phi) v}) &\ge f(\phi \sqrt{u} + (1-\phi) \sqrt{v})\intertext{since $f$ is monotonically increasing and $\sqrt{\cdot}$ is concave.}
	&\ge \phi f(\sqrt{u}) + (1-\phi) f(\sqrt{v})
	\end{align*}
	since $f$ is concave on $\R_{\ge 0}$. This can be shown by considering
	\[\frac{d^2}{du^2} f(u) = \frac{e^{u}}{(1+e^{u})^3}(1-e^{u}) \le 0 \; \forall u \ge 0\]
	Combining these facts, we have
	\begin{align*}
	\E_{p_L}[f(L)] &\le \E_{p_L}[f(\abs{L})]&\text{since $f(\ell) \le f(\abs{\ell})$}
	\\&= \E_{p_L}\bigl[f\bigl(\sqrt{L^2}\bigr)\bigr]
	\\&\le f\Bigl(\sqrt{\E_{p_L}[L^2]}\Bigr)&\text{from Jensen's inequality with the concavity of $f(\sqrt{\cdot})$}
	\\&\le f(\sqrt{P})
	\end{align*}
	since $f(\sqrt{\cdot})$ is monotonically increasing, and by assumption $E_{p_L}[L^2] \le P$. Similarly, $\E_{p_L}[1 - f(L)] \le f(\sqrt{P})$, and therefore we can set $\varepsilon = 1 - f(\sqrt{P})$. Applying this choice of $\varepsilon$ to \eqref{eq:W2upper-eps} we have
	\[\abs{I(p_L,f) - I(B_t,f)}  \le \biggl(K_1 \log_2\biggl(\frac{f(\sqrt{P})}{1-f(\sqrt{P})}\biggr) + K_2 \biggr) W_2(p_L,B_t)\]
	and can set $K_P = K_1 \log_2\biggl(\frac{f(\sqrt{P})}{1-f(\sqrt{P})}\biggr) + K_2$ to obtain $\abs{I(p_L,f) - I(B_t,f)} \le K_P W_2(p_L,B_t)$.
	
	Recall that $C = \max_{p_L \in \mathcal{C}_P} I(p_L, f)= I(B_{\sqrt{P}}, f)$ and $\widetilde{C}_n = \max_{x \in \Ucal_n} I(p_{L_n \mid \Lcal_{n-1}}, f)$. By assumption, $P$ is selected such that $p_{L_n \mid \Lcal_{n-1}} \in \mathcal{C}_P$ for any $x \in \Ucal_n$, which implies $I(p_{L_n \mid \Lcal_{n-1}},f) \le C$ for any $x \in \Ucal_n$ and hence $\widetilde{C}_n \le C$. Combining these facts, we have
	\[\widetilde{C}_n - I(p_{L_n \mid \Lcal_{n-1}}, f) \le C - I(p_{L_n \mid \Lcal_{n-1}}, f) = \abs{I(B_{\sqrt{P}},f) - I(p_{L_n \mid \Lcal_{n-1}}, f)} \le K_P W_2(p_{L_n \mid \Lcal_{n-1}},B_{\sqrt{P}}).\qedhere\]
\end{proof}

\subsection{Proof of Proposition 3.2}
\begin{proof}
	Adopting notation from \citet{merigot2011}, let $S$ denote a finite set of points in $\R$, and $w\colon S \to \R$ a weight vector. Define $\operatorname{Vor}_S^w(p) = \{\ell : \norm{\ell - p}_2^2 - w(p) \le \norm{\ell - q}_2^2 - w(q) \enspace \forall q \in S\}$.
	
	Let $\mu$ be a given probability measure with density $p_L$. Consider $S = \{-t,t\}$, with the corresponding measure $B_t = \sum_{p \in S} \frac12 \delta_p = \frac12 \delta_{-t} + \frac12 \delta_{t}$. Let $w^*(-t) = 2t \med_{p_L}(L)$, and $w^*(t) = -2t \med_{p_L}(t)$. We have
	\begin{align*}
	\operatorname{Vor}_S^{w^*}(-t) &= \{\ell : \norm{\ell + t}_2^2 - w^*(-t) \le \norm{\ell - q}_2^2 - w^*(q) \enspace \forall q \in \{-t,t\}\}
	\\ &= \{\ell : \norm{\ell + t}_2^2 - w^*(-t) \le \norm{\ell - t}_2^2 - w^*(t)\}
	\\ &= \{\ell : \norm{\ell + t}_2^2 - 2t \med_{p_L}(L) \le \norm{\ell - t}_2^2 + 2t \med_{p_L}(L)\}
	\\ &= \{\ell : \ell \le  \med_{p_L}(L)\}
	\end{align*}
	and similarly $\operatorname{Vor}_S^{w^*}(t)=\{\ell : \ell \ge  \med_{p_L}(L)\}$.
	We have
	\[\int_{\operatorname{Vor}_S^{w^*}(-t)} p_L(\ell) d\ell=\int_{\ell \le  \med_{p_L}(L)} p_L(\ell) d\ell=\frac12\]
	and similarly $\int_{\operatorname{Vor}_S^{w^*}(t)} p_L(\ell) d\ell=\frac12$. Therefore, $w^*$ is \emph{adapted} to $(\mu,B_t)$. By Theorem 2 of \citet{merigot2011}, a map $T_S^{w^*} \colon \R \to \R$ exists which realizes an optimal transport between $\mu$ and $B_t$. By \citet{merigot2011} Theorem 1, we have
	\begin{align*}
	W_2^2(\mu,B_t) &= \int_{\operatorname{Vor}_S^{w^*}(-t)} \norm{\ell + t}_2^2\;p_L(\ell) d\ell + \int_{\operatorname{Vor}_S^{w^*}(t)} \norm{\ell - t}_2^2\;p_L(\ell) d\ell
	\\ &= \int_{\ell \le  \med_{p_L}(L)} \norm{\ell + t}_2^2\;p_L(\ell) d\ell + \int_{\ell \ge \med_{p_L}(L)} \norm{\ell - t}_2^2\;p_L(\ell) d\ell
	\\ &= \E_{p_L}[L^2] + t^2 - 2t \biggl(\int_{\ell \ge  \med_{p_L}(L)} \ell\;p_L(\ell) d\ell - \int_{\ell \le \med_{p_L}(L)} \ell\;p_L(\ell) d\ell \biggr)
	\\ &= \E_{p_L}[L^2] + t^2 - 2t \biggl(\int_{\ell \ge  \med_{p_L}(L)} (\ell-\med_{p_L}(L))\;p_L(\ell) d\ell + \int_{\ell \le \med_{p_L}(L)} (\med_{p_L}(L) - \ell)\;p_L(\ell) d\ell\biggr)
	\\ &= \E_{p_L}[L^2] + t^2 - 2t \biggl(\int_{\ell \ge  \med_{p_L}(L)} \abs{\ell-\med_{p_L}(L)}\;p_L(\ell) d\ell + \int_{\ell \le \med_{p_L}(L)} \abs{\ell - \med_{p_L}(L)}\;p_L(\ell) d\ell\biggr)
	\\ &= \E_{p_L}[L^2] + t^2 - 2t \E_{p_L}[\abs{L - \med_{p_L}(L)}]\qedhere
	\end{align*}
\end{proof}
\subsection{Proof of Corollary 3.2.1}
\begin{proof}
	Let $p_L \sim \mathcal{N}(\mu,\sigma^2)$. We have $\E_{p_L}[L^2] = \E_{p_L}[L]^2 + \operatorname{Var}_{p_L}(L) = \mu^2 + \sigma^2$, and $\E_{p_L}[\abs{L - \med_{p_L}(L)}] = \E_{p_L}[\abs{L - \mu}] = \sigma \sqrt{\frac{2}{\pi}}$ \citep{winkelbauer2014moments}. Hence $W_2^2(p_L,B_t) = \E_{p_L}[L^2] + t^2 - 2t \E_{p_L}[\abs{L - \med_{p_L}(L)}] = \mu^2 + \sigma^2 + t^2 - 2 \sqrt{\frac{2}{\pi}} t \sigma$. Completing the square, we have the desired result.
\end{proof}

\section{Experiment Details}
\label{app:exp}

\subsection{Selection of Power Constraint}\label{app:power}
Recall that APM-LR minimizes an objective function consisting of a mixture of two terms, reprinted below:
\begin{equation}
\pi_n(\Lcal_{n-1}) = \argmin_{x \in \Ucal_n}\enspace(\mu_n^T x)^2 + \biggl(\sqrt{x^T \Sigma_n x}-\sqrt{\frac{2}{\pi}P_n}\biggr)^2.\label{eq:APMLR-supp-power}
\end{equation}
The first term in \eqref{eq:APMLR-supp-power}, which is independent of $P_n$, encourages $x$ to lie orthogonal to the hyperplane posterior mean, $\mu_n$. For all such $x$ satisfying $\mu_n^T x = 0$, we have $\E[L_n] = \mu_n^T x = 0$ and
\[\E[L_n^2] = (\mu_n^T x)^2 + x^T \Sigma_n x = x^T \Sigma_n x \le B^2 \lambda_1(\Sigma_n)\]
where expectations are taken with respect to $p_{L_n \mid \Lcal_{n-1}}$. Therefore $P_n = B^2 \lambda_1(\Sigma_n)$ is a valid power constraint for the set of examples that induce zero-mean input distributions. This set arguably contains the ``best'' candidate examples, since if $(\mu_n^T x)^2 \gg 0$ then the objective in \eqref{eq:APMLR-supp-power} will be large. For this reason we set $P_n = B^2 \lambda_1(\Sigma_n)$ in our experiments, as opposed to the power constraint of $B^2 \lambda_1(\mu_n \mu_n^T + \Sigma_n)$ which is valid for all examples but is loose for examples encouraged by the first term in \eqref{eq:APMLR-supp-power}.

\clearpage 
\subsection{Dataset Information}\label{app:data}
In Table \ref{tab:datasets-full} we describe the datasets used in our experiments. Several datasets have multiple classes: in this case, we select a two-class dataset partition by either grouping individual classes together into super-classes, or simply training on a subset of the classes. In our experiments we treat each class partition as its own dataset, and refer to each partition by a nickname. All datasets except for \emph{clouds}, \emph{cross}, and \emph{horseshoe} come from the UCI Machine Learning Repository \citep{dua2017}; several UCI datasets have additional citations, which are listed next to their names.

\begin{table}[h]
	\renewcommand{\arraystretch}{1.2}
	\centering
	\begin{small}
		\begin{tabular}{|M{18ex}|M{25ex}|L{25ex}|M{9ex}|M{10ex}|}
			\hline
			Nickname & Dataset & \centering{Class partition} &  \# of features & \# of examples\\
			\hline\hline
			\emph{vehicle-full} & Vehicle Silhouettes \citep{enlighten91397} & $Y=-1$: `saab' or `opel' \newline $Y=1$:\hspace{11pt}`bus' or `van' & 18 & 846\\
			\hline
			\emph{vehicle-cars} & Vehicle Silhouettes \citep{enlighten91397}& $Y=-1$: `saab' \newline $Y=1$:\hspace{7pt} `opel' & 18 & 429\\
			\hline
			\emph{vehicle-transport} & 	Vehicle Silhouettes \citep{enlighten91397}& $Y=-1$: `bus' \newline $Y=1$:\hspace{7pt} `van' & 18 & 417\\
			\hline
			\emph{letterDP} & Letter Recognition &  $Y=-1$: `D' \newline $Y=1$:\hspace{7pt} `P' & 16 & 1608 \\
			\hline
			\emph{letterEF} & Letter Recognition &  $Y=-1$: `E' \newline $Y=1$:\hspace{7pt} `F' & 16 & 1543 \\
			\hline
			\emph{letterIJ} & Letter Recognition &  $Y=-1$: `I' \newline $Y=1$:\hspace{7pt} `J' & 16 & 1502 \\
			\hline
			\emph{letterMN} & Letter Recognition &  $Y=-1$: `M' \newline $Y=1$:\hspace{7pt} `N' & 16 & 1575 \\
			\hline
			\emph{letterUV} & Letter Recognition &  $Y=-1$: `U' \newline $Y=1$:\hspace{7pt} `V' & 16 & 1577 \\
			\hline
			\emph{letterVY} & Letter Recognition & $Y=-1$: `V' \newline $Y=1$:\hspace{7pt} `Y' & 16 & 1550 \\
			\hline
			\emph{austra} & Australian Credit Approval &  $Y=-1$: `0' \newline $Y=1$:\hspace{7pt} `1' & 14 & 690 \\
			\hline
			\emph{wdbc} & Breast Cancer Wisconsin (Diagnostic) & $Y=-1$: `M' \newline $Y=1$:\hspace{7pt} `B' & 30 & 569 \\
			\hline
			\emph{clouds} & \hspace{15pt}Synth1\newline\citep{yang2018} & $Y=-1$: `-1' \newline $Y=1$:\hspace{7pt} `1' & 2 & 600 \\
			\hline
			\emph{cross} & \hspace{15pt}Synth2\newline\citep{yang2018} &  $Y=-1$: `-1' \newline $Y=1$:\hspace{7pt} `1' & 2 & 600 \\
			\hline
			\emph{horseshoe} & \hspace{15pt}Synth3\newline\citep{yang2018} & $Y=-1$: `-1' \newline $Y=1$:\hspace{7pt} `1' & 2 & 600 \\
			\hline
		\end{tabular}
	\end{small}
	\caption{Full dataset information}
	\label{tab:datasets-full}
\end{table}

\clearpage
\subsection{Baseline Methods Details}\label{app:methods}
Below we elaborate on the BALD and InfoGain baseline selection methods:

\paragraph{InfoGain} We can directly approximate information gain $I(\theta;Y \mid \Lcal_{n-1})$ with a Monte Carlo approximation over $s$ samples from $p_{\theta \mid \Lcal_{n-1}} \sim \mathcal{N}(\mu_n,\Sigma_n)$:
\begin{align}
I(\theta;Y \mid \Lcal_{n-1}) &= h_b(\E_{p_{\theta \mid \Lcal_{n-1}}}[f(\theta^T x_n)]) - \E_{p_{\theta \mid \Lcal_{n-1}}}[h_b(f(\theta^T x_n))]\notag
\\ &\approx h_b\biggl(\frac1s \sum_{i=1}^s f(\theta_i^T x_n)\biggr) - \frac1s \sum_{i=1}^s h_b\biggl(f(\theta_i^T x_n)\biggr) \qquad \theta_i \sim p_{\theta \mid \Lcal_{n-1}}\notag
\\ &\approx h_b\biggl(\frac1s \sum_{i=1}^s f(\theta_i^T x_n)\biggr) - \frac1s \sum_{i=1}^s h_b\biggl(f(\theta_i^T x_n)\biggr) \qquad \theta_i \sim \mathcal{N}(\mu_n,\Sigma_n)\label{eq:infogainMC}
\end{align}
Our ``InfoGain'' baseline selects the example $x_n \in \Ucal_n$ that maximizes the expression in \eqref{eq:infogainMC}, computed in $O(sd)$ time per candidate example.

\paragraph{BALD} Consider a probit regression label distribution $p(Y=1 \mid L) = \Phi(L)$, where $\Phi$ is the standard normal cumulative distribution function. For $p_L \sim \mathcal{N}(\mu,\sigma^2)$, \citet{houlsby2011bayesian} use a Taylor expansion in the BALD algorithm to approximate $I(p_L,\Phi(L))$ as
\begin{equation}
I(p_L,\Phi(L)) \approx h_b\biggl(\Phi\biggl(\frac{\mu}{\sqrt{\sigma^2 + 1}}\biggr)\biggr)-\frac{D\operatorname{exp}\Bigl(-\frac{\mu^2}{2(\sigma^2 + D^2)}\Bigr)}{\sqrt{\sigma^2 + D^2}}\label{eq:BALDraw}
\end{equation}
where $D = \sqrt{\frac{\pi \ln2}{2}}$. By equalizing derivatives at $L=0$, we can approximate $f(L) \approx \Phi(kL)$ where $k=\sqrt{\frac{\pi}{8}}$ \citep{bishop2006}. Define $\widetilde{L}=kL$ and note that $\widetilde{L} \sim \mathcal{N}(\widetilde{\mu}, \widetilde{\sigma}^2)$ for $\widetilde{\mu}=k\mu$ and $\widetilde{\sigma}^2=k^2\sigma^2$. We can then use the BALD approximation in \eqref{eq:BALDraw} for logistic regression:
\begin{align*}
I(p_L,f(L)) &\approx I(p_L,\Phi(kL))
\\ &=h_b(\E_{p_L}(\Phi(kL)))-\E_{p_L}(h_b(\Phi(kL)))
\\ &=h_b(\E_{p_{\widetilde{L}}}(\Phi(\widetilde{L})))-\E_{p_{\widetilde{L}}}(h_b(\Phi(\widetilde{L})))
\\ &= I(p_{\widetilde{L}},\Phi(\widetilde{L}))
\\ &\approx h_b\Biggl(\Phi\Biggl(\frac{k\mu}{\sqrt{k^2\sigma^2 + 1}}\Biggr)\Biggr)-\frac{D\operatorname{exp}\Bigl(-\frac{k^2\mu^2}{2(k^2\sigma^2 + D^2)}\Bigr)}{\sqrt{k^2\sigma^2 + D^2}}
\end{align*}

Approximating $p_{\theta \mid \Lcal_{n-1}} \sim \mathcal{N}(\mu_n,\Sigma_n)$, we have $p_{L_n \mid \Lcal_{n-1}} \sim \mathcal{N}(\mu_n^T x_n, x_n^T \Sigma_n x_n)$ and so we can approximate
\begin{equation}
I(p_{L_n \mid \Lcal_{n-1}},f(L))\approx h_b\Biggl(\Phi\Biggl(\frac{k\mu_n^T x_n}{\sqrt{k^2 x_n^T \Sigma_n x_n + 1}}\Biggr)\Biggr)-\frac{D\operatorname{exp}\Bigl(-\frac{k^2 (\mu_n^T x_n)^2}{2(k^2 x_n^T \Sigma_n x_n + D^2)}\Bigr)}{\sqrt{k^2 x_n^T \Sigma_n x_n + D^2}}\label{eq:BALDfinal}
\end{equation}
where $D = \sqrt{\frac{\pi \ln2}{2}}$ and $k=\sqrt{\frac{\pi}{8}}$. Our ``BALD'' baseline method selects the example $x_n \in \Ucal_n$ that maximizes the expression in \eqref{eq:BALDfinal}, computed in $O(d^2)$ time per candidate example.

\paragraph{Summary} For completeness, below we summarize all selection methods used in our experiments. For any method utilizing a normal approximation to the hyperplane posterior, let $p_{\theta \mid \Lcal_{n-1}} \sim \mathcal{N}(\mu_n,\Sigma_n)$. Let $\widehat{\theta}_{n-1} = \A(\Lcal_{n-1})$, $D = \sqrt{\frac{\pi \ln2}{2}}$, and $k=\sqrt{\frac{\pi}{8}}$.
\begin{align}
\text{\emph{APM-LR}:}&\quad x_n = \argmin_{x \in \Ucal_n}\enspace(\mu_n^T x)^2 + \biggl(\sqrt{x^T \Sigma_n x}-\sqrt{\frac{2}{\pi}P_n}\biggr)^2 \label{eq:apmlr-supp}\\
\text{\emph{Uncertainty}:}&\quad x_n = \argmin_{x \in \Ucal_n}\enspace x^T \widehat{\theta}_{n-1} \notag\\
\text{\emph{Random}:}&\quad \text{Select $x_n$ uniformly at random from $\Ucal_n$}\notag\\
\text{\emph{MaxVar}:}&\quad x_n = \argmax_{x \in \Ucal_n}\enspace x^T \Sigma_n x \notag\\
\text{\emph{InfoGain}:}&\quad x_n = \argmax_{x \in \Ucal_n}\enspace h_b\biggl(\frac1s \sum_{i=1}^s f(\theta_i^T x_n)\biggr) - \frac1s \sum_{i=1}^s h_b\biggl(f(\theta_i^T x_n)\biggr) \qquad \theta_i \sim \mathcal{N}(\mu_n,\Sigma_n) \notag\\
\text{\emph{BALD}:}&\quad x_n = \argmax_{x \in \Ucal_n}\enspace h_b\Biggl(\Phi\Biggl(\frac{k\mu_n^T x_n}{\sqrt{k^2 x_n^T \Sigma_n x_n + 1}}\Biggr)\Biggr)-\frac{D\operatorname{exp}\Bigl(-\frac{k^2 (\mu_n^T x_n)^2}{2(k^2 x_n^T \Sigma_n x_n + D^2)}\Bigr)}{\sqrt{k^2 x_n^T \Sigma_n x_n + D^2}} \notag
\end{align}

\subsection{Extended Test Accuracy Results}
\label{app:test}
Below we plot average holdout test accuracy against number of queried examples, excluding one initial seed point selected uniformly at random per class. Error bars show $\pm1$ standard error over 150 trials per method. For visual clarity, we display different numbers of queried examples for each dataset.

\begin{figure}[h]
	\centering
	\begin{subfigure}[t]{0.33\textwidth}
		\centering
		\includegraphics[width=\textwidth]{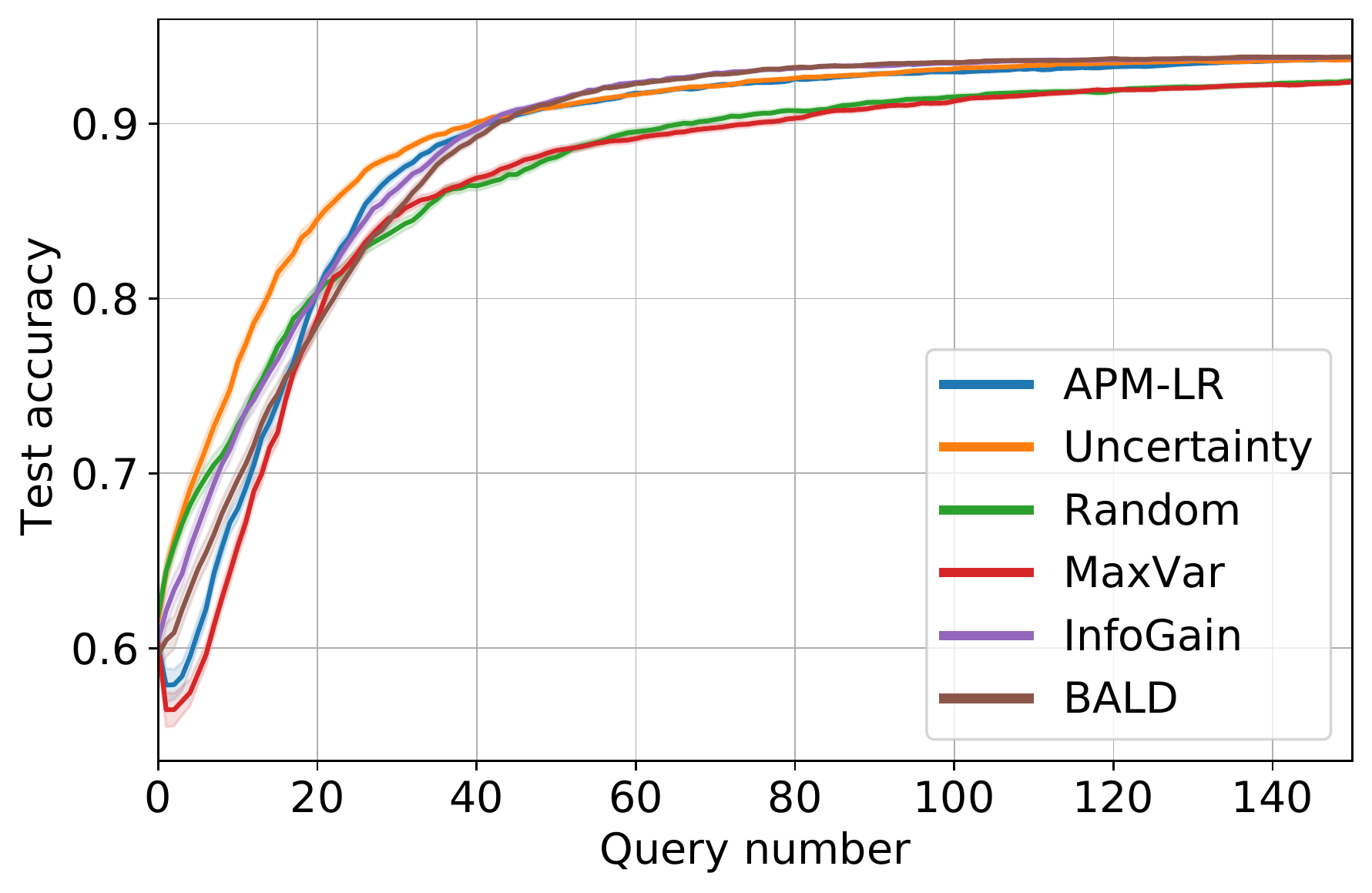}
		\caption{\emph{vehicle-full}}
		\label{fig:vehicle-full}
	\end{subfigure}%
	~
	\begin{subfigure}[t]{0.33\textwidth}
		\centering
		\includegraphics[width=\textwidth]{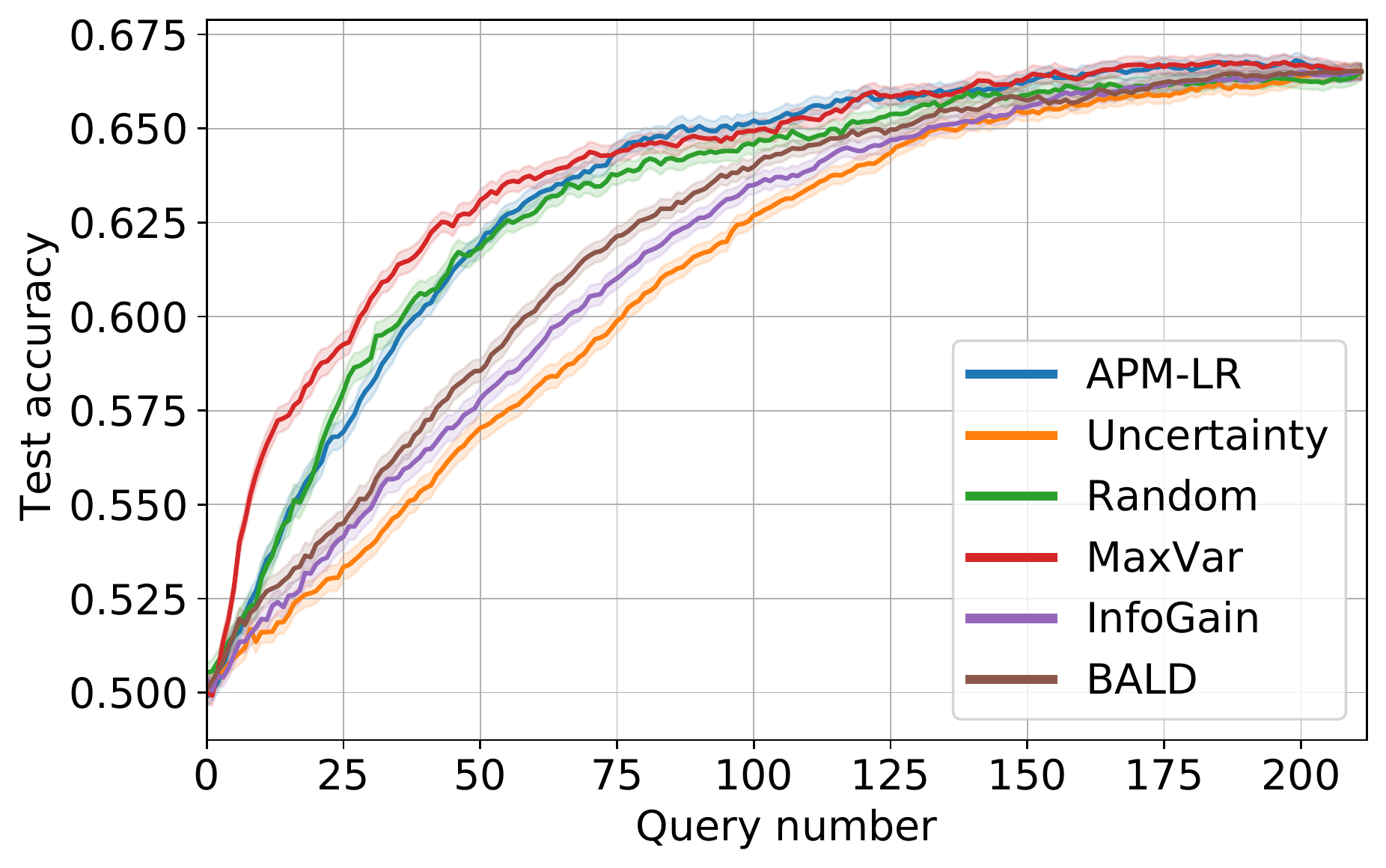}
		\caption{\emph{vehicle-cars}}
		\label{fig:vehicle-cars}
	\end{subfigure}%
	~
	\begin{subfigure}[t]{0.33\textwidth}
		\centering
		\includegraphics[width=\textwidth]{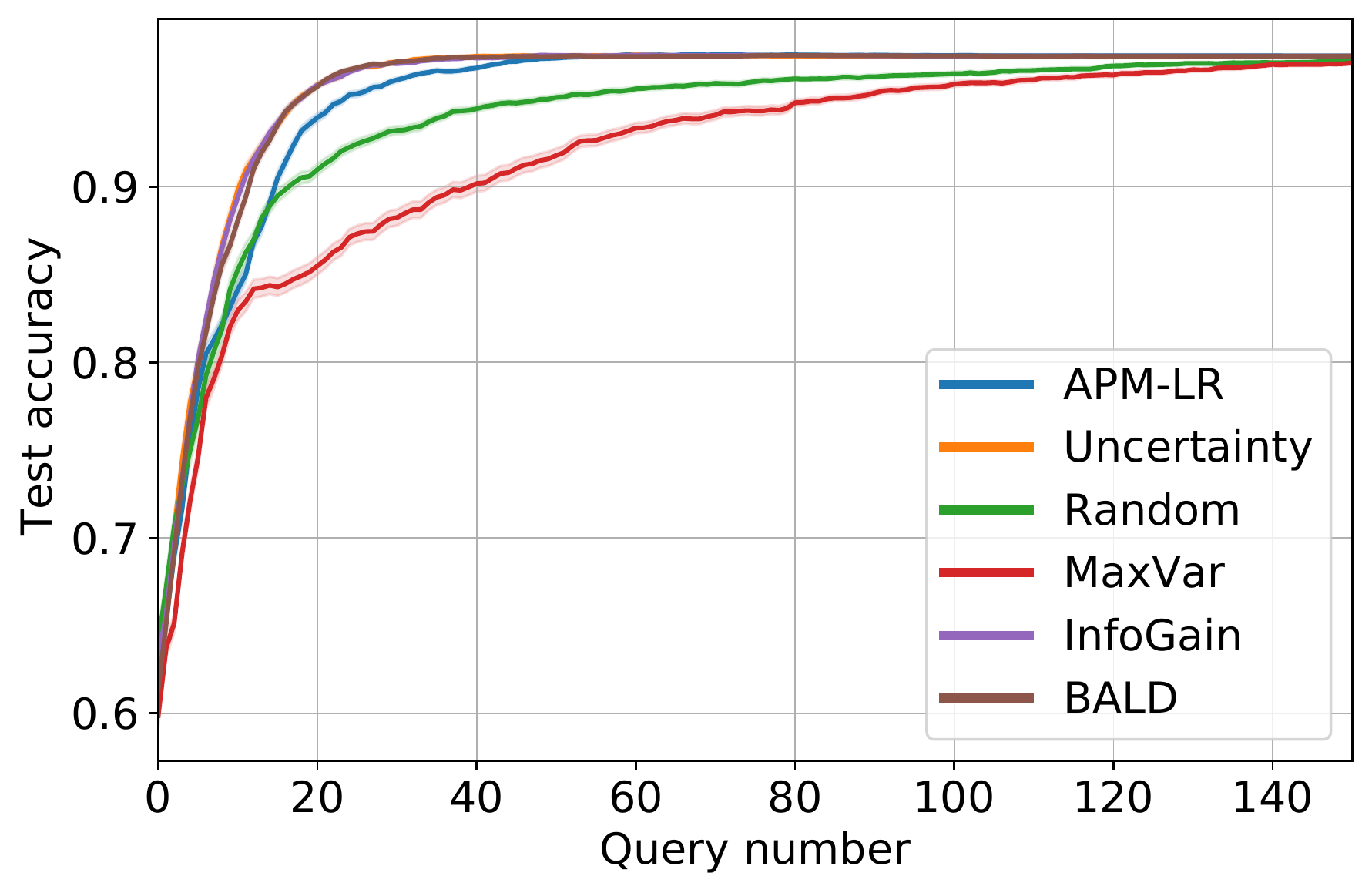}
		\caption{\emph{vehicle-transport}}
		\label{fig:vehicle-transport}
	\end{subfigure}%
	\caption{Test accuracy on ``Vehicle Silhouettes''}
	\label{fig:vehicle}
\end{figure}

Figure \ref{fig:vehicle} shows test accuracy across several two-class partitions of the Vehicle Silhouettes dataset (see Table \ref{tab:datasets-full}). In \emph{vehicle-cars}, Uncertainty, InfoGain, and BALD fail to perform as well as MaxVar, Random, and APM-LR. As noted in \citet{yang2018}, there are cases where Random sampling --- or more generally, selection methods that encourage dataset exploration --- can outperform methods that maximize information. In \emph{vehicle-cars}, it's possible that the ``exploration'' component in APM-LR encourages the selection of satisfactory examples, which we investigate further in Section \ref{app:fail}.

\begin{figure}[h]
	\centering
	\begin{subfigure}[t]{0.33\textwidth}
		\centering
		\includegraphics[width=\textwidth]{fullrun_letterDP}
		\caption{\emph{letterDP}}
		\label{fig:letter-DvP}
	\end{subfigure}%
	~
	\begin{subfigure}[t]{0.33\textwidth}
		\centering
		\includegraphics[width=\textwidth]{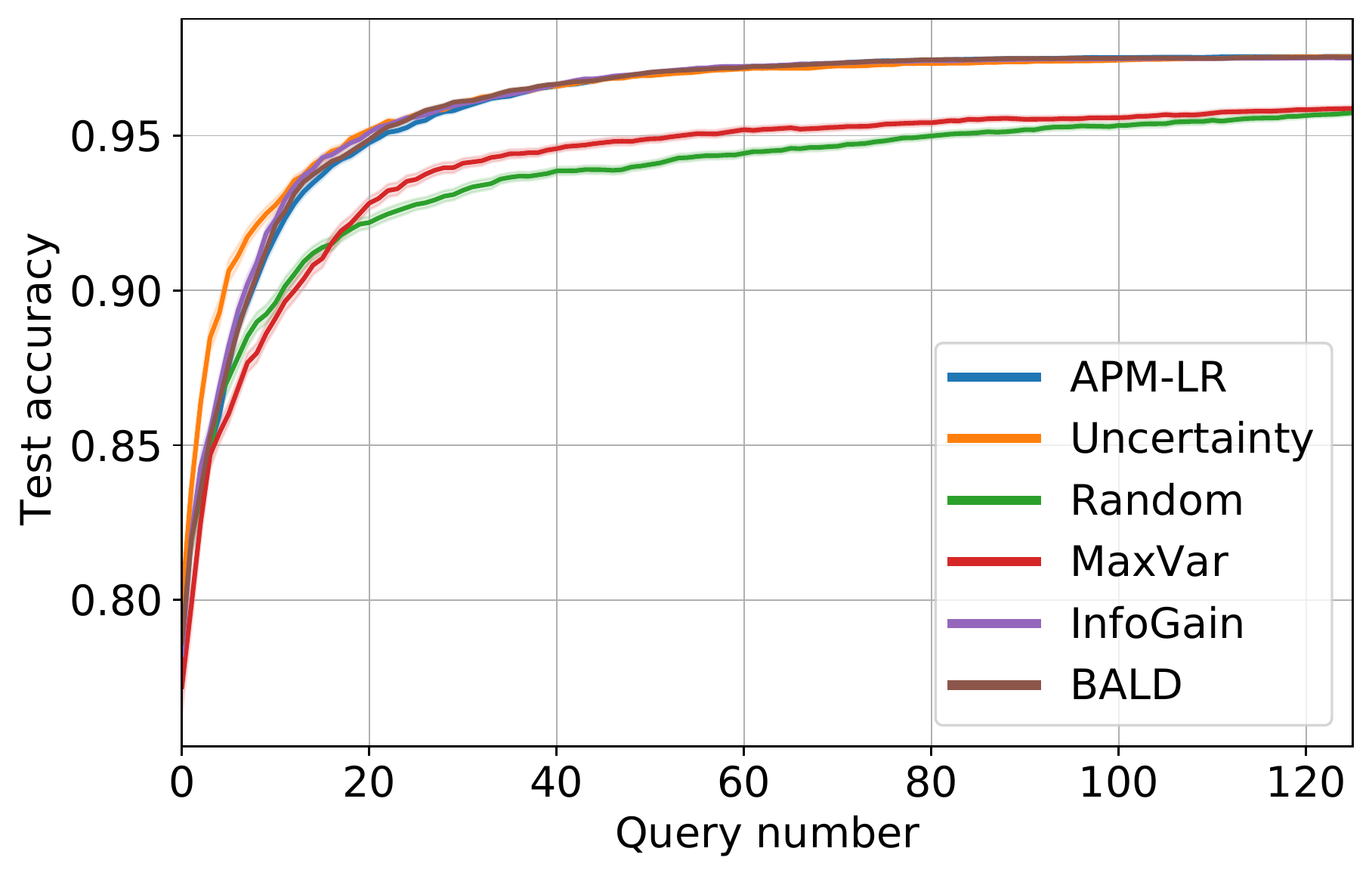}
		\caption{\emph{letterEF}}
		\label{fig:letter-EvF}
	\end{subfigure}%
	~
	\begin{subfigure}[t]{0.33\textwidth}
		\centering
		\includegraphics[width=\textwidth]{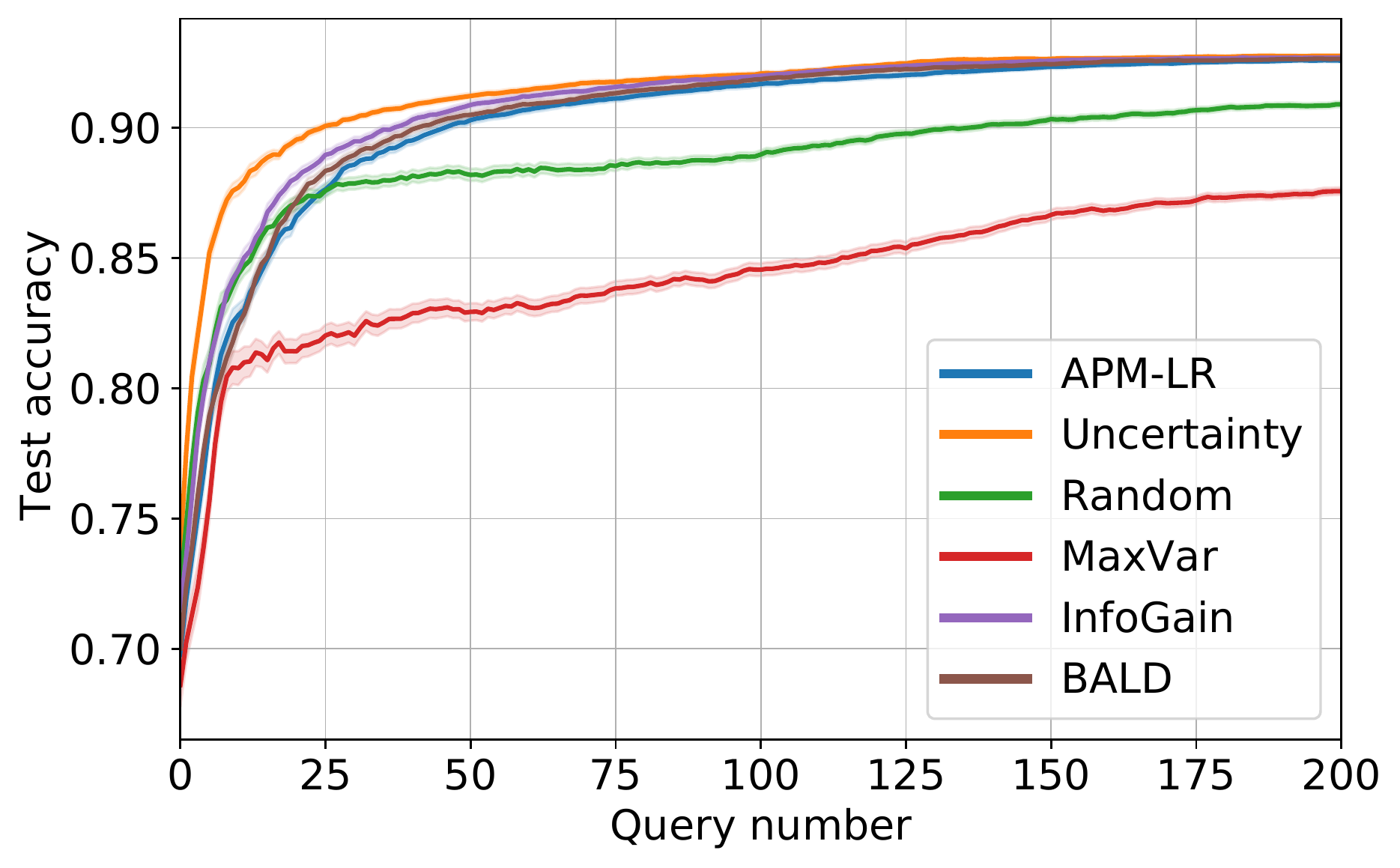}
		\caption{\emph{letterIJ}}
		\label{fig:letter-IvJ}
	\end{subfigure}%
	\\
	\begin{subfigure}[t]{0.33\textwidth}
		\centering
		\includegraphics[width=\textwidth]{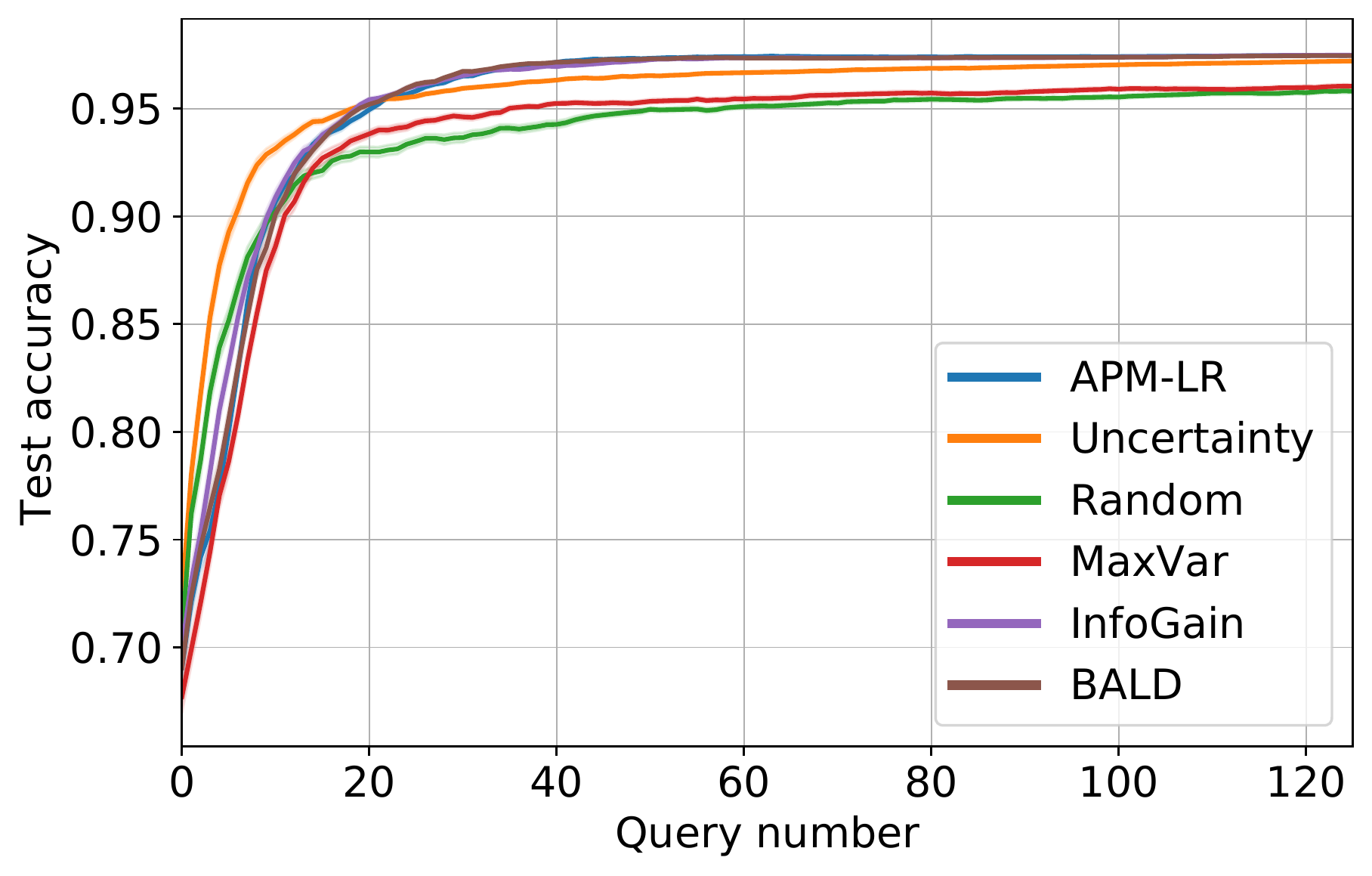}
		\caption{\emph{letterMN}}
		\label{fig:letter-MvN}
	\end{subfigure}%
	~
	\begin{subfigure}[t]{0.33\textwidth}
		\centering
		\includegraphics[width=\textwidth]{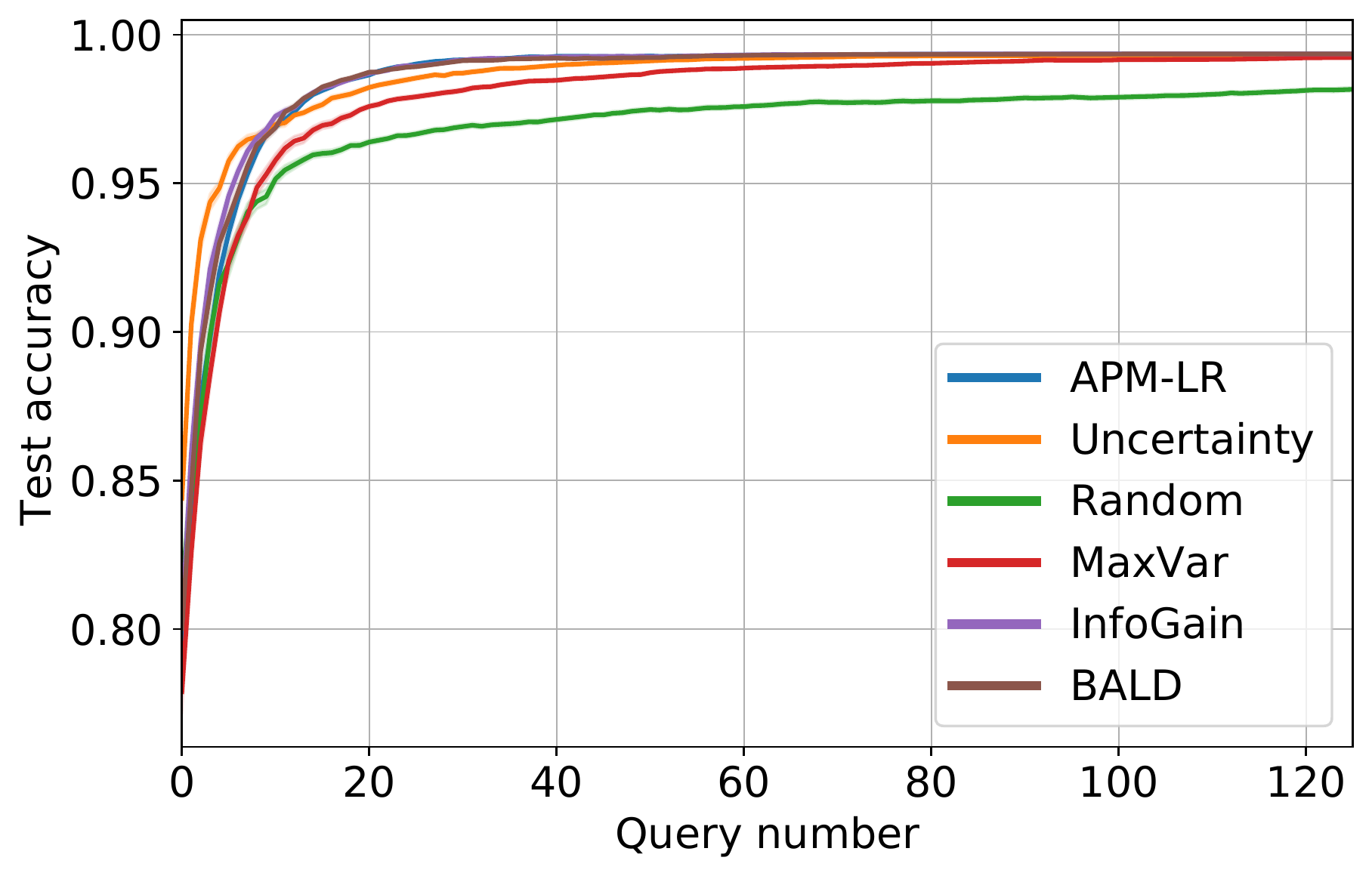}
		\caption{\emph{letterUV}}
		\label{fig:letter-UvV}
	\end{subfigure}%
	~
	\begin{subfigure}[t]{0.33\textwidth}
		\centering
		\includegraphics[width=\textwidth]{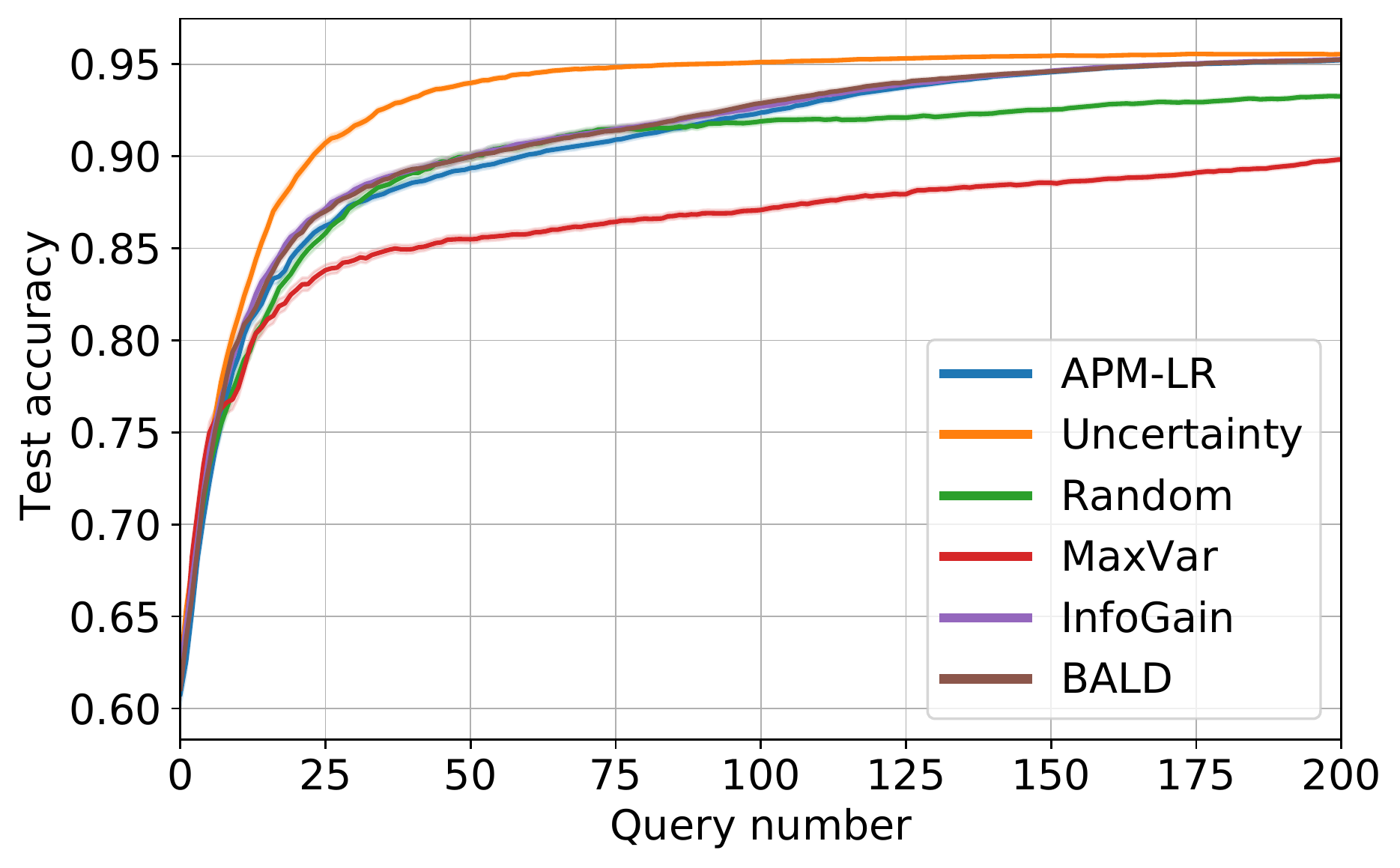}
		\caption{\emph{letterVY}}
		\label{fig:letter-VvY}
	\end{subfigure}%
	\caption{Test accuracy on ``Letter Recognition''}
	\label{fig:letter}
\end{figure}

\clearpage
Figure \ref{fig:letter} shows test accuracy across several two-class partitions of the Letter Recognition dataset. All partitions show similar trends to \emph{letterDP}, which was included in the paper body.

\begin{figure}[h]
	\centering
	\begin{subfigure}[t]{0.33\textwidth}
		\centering
		\includegraphics[width=\textwidth]{fullrun_austra}
		\caption{\emph{austra}}
		\label{fig:australian}
	\end{subfigure}%
	~
	\begin{subfigure}[t]{0.33\textwidth}
		\centering
		\includegraphics[width=\textwidth]{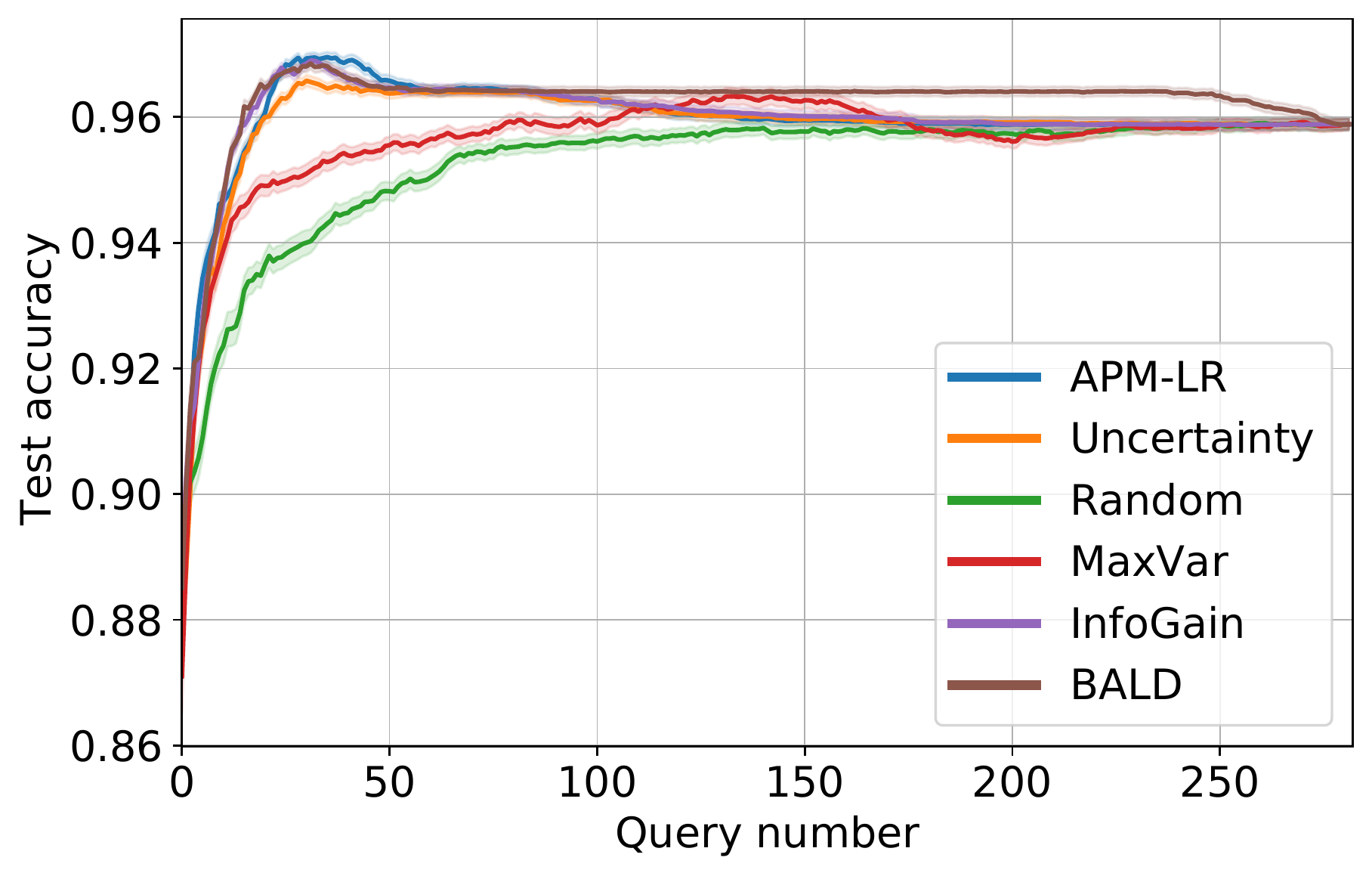}
		\caption{\emph{wdbc}}
		\label{fig:wdbc}
	\end{subfigure}%
	\caption{Miscellaneous UCI datasets}
	\label{fig:misc}
\end{figure}

Figure \ref{fig:misc} shows test accuracy across the remaining UCI datasets in Table \ref{tab:datasets-full}. On \emph{wdbc}, the active methods appear to have an average test accuracy that peaks early and then gradually decreases. While this behavior merits further investigation, we note that it is possible in some cases for a selected subset of the full data pool to generalize better than when training on the entire pool \citep{ma2018teacher}.

\begin{figure}[h]
	\centering
	\begin{subfigure}[t]{0.33\textwidth}
		\centering
		\includegraphics[width=\textwidth]{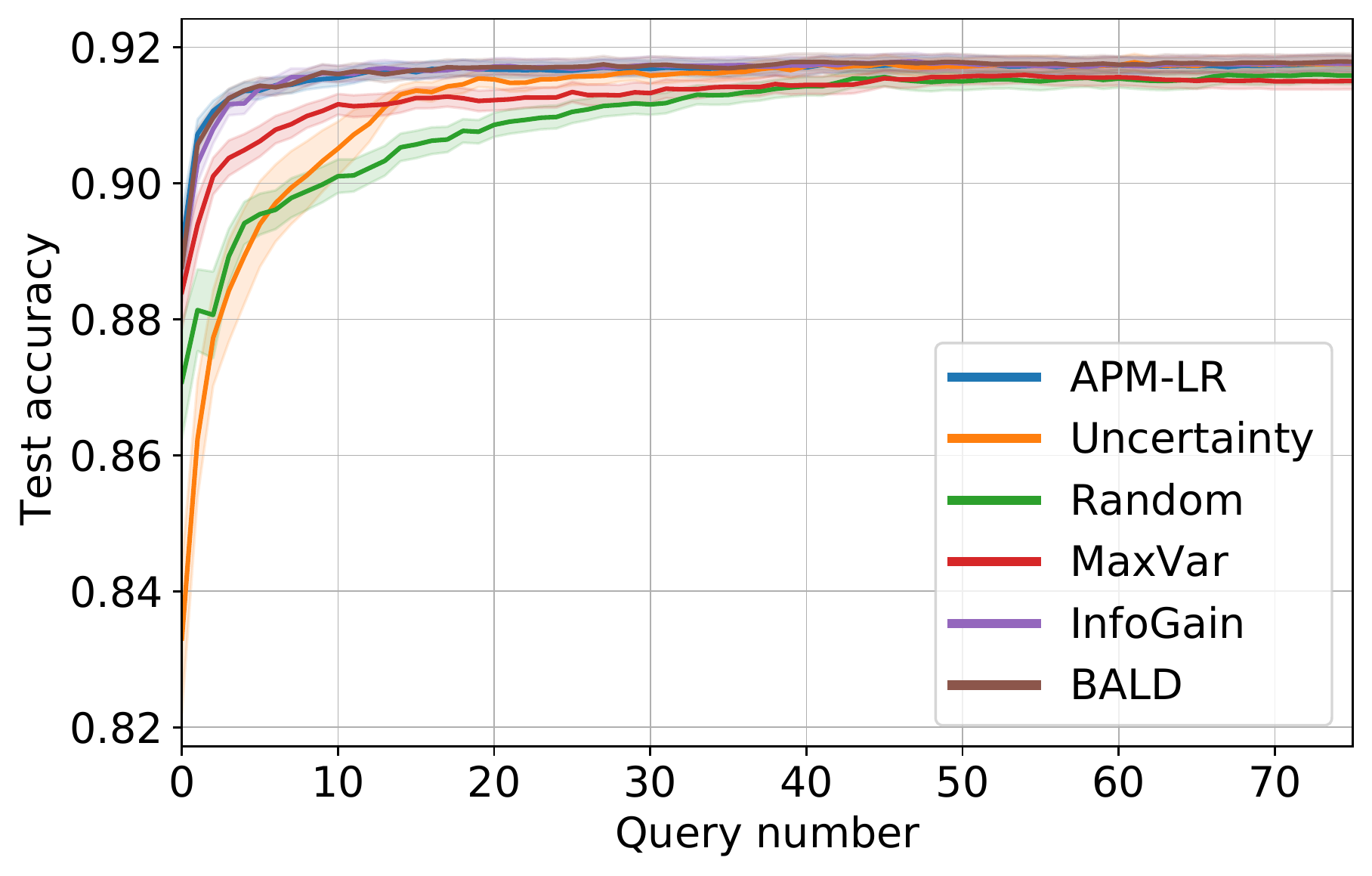}
		\caption{\emph{clouds}}
		\label{fig:CLOUDS}
	\end{subfigure}%
	~
	\begin{subfigure}[t]{0.33\textwidth}
		\centering
		\includegraphics[width=\textwidth]{fullrun_cross}
		\caption{\emph{cross}}
		\label{fig:CROSS}
	\end{subfigure}%
	~
	\begin{subfigure}[t]{0.33\textwidth}
		\centering
		\includegraphics[width=\textwidth]{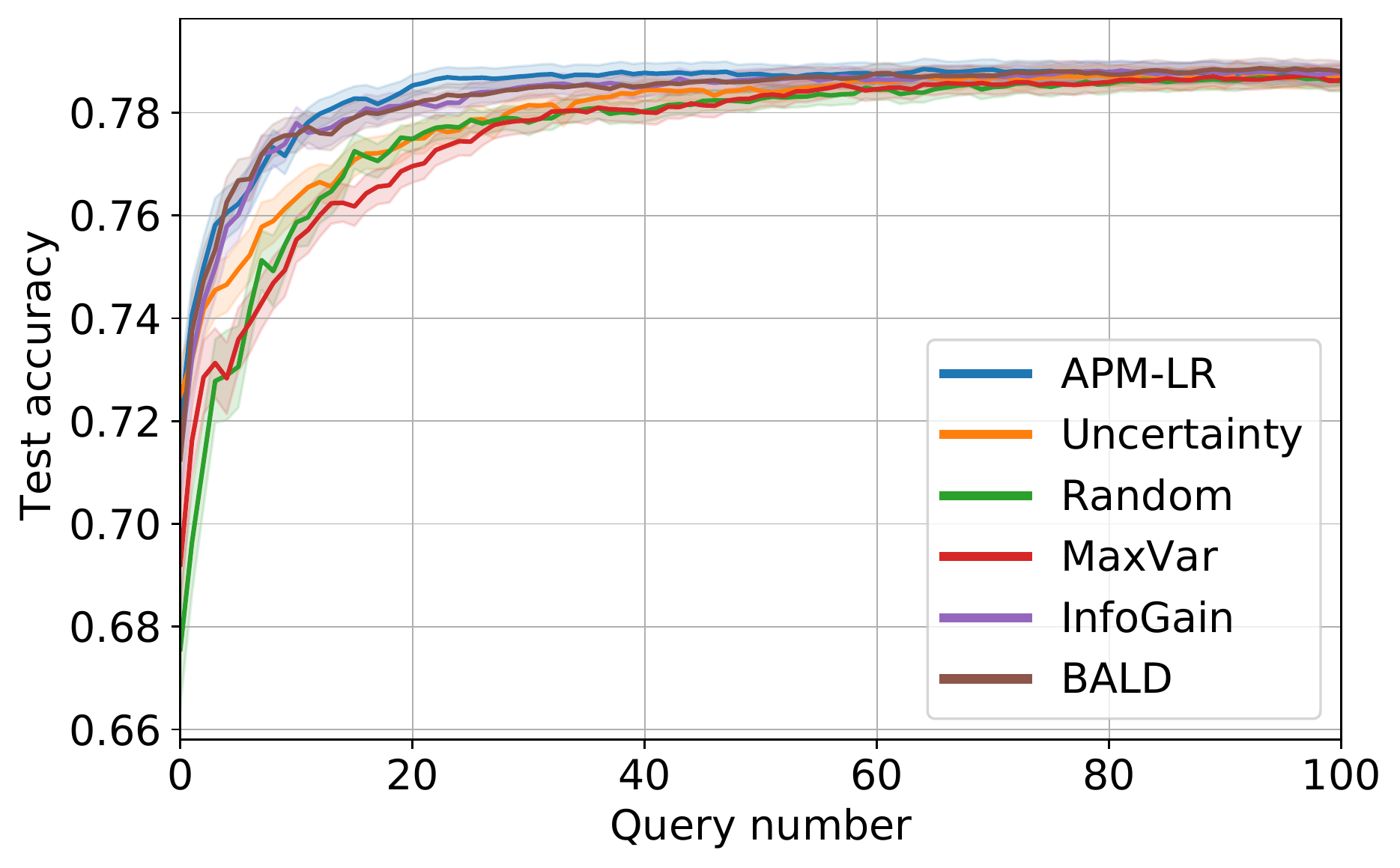}
		\caption{\emph{horseshoe}}
		\label{fig:HORSESHOE}
	\end{subfigure}%
	\caption{Synthetic datasets}
	\label{fig:synthetic}
\end{figure}

\clearpage
Figure \ref{fig:synthetic} shows test accuracy across several synthetic datasets. On \emph{clouds} and \emph{cross}, Uncertainty sampling is outperformed by the other baseline active learning methods, except MaxVar.
\subsection{Extended Computational Cost Results}\label{app:comp}
All experiments were run on Intel Xeon Gold 6226 CPUs at 2.7 GHz. In Table \ref{tab:selection_cost} we present for all datasets the cumulative compute time (in seconds) needed for each method to select the first 40 examples (excluding seed points). In this first table, we exclude the compute time needed to retrain the logistic regression model and perform the VariationalEM posterior update after each example is selected, since these steps are common to all selection methods. While some methods do not directly utilize the variational posterior in selecting examples, we perform variational posterior updates for all data selection methods since we consider the variational posterior to be part of the Bayesian model produced by the training routine.
\begin{table}[h]
	\centering
	\begin{small}
\begin{tabular}{|S[table-format=2.3]|S[table-format=2.3]|S[table-format=2.3]|S[table-format=2.3]|S[table-format=2.3]|S[table-format=2.3]|S[table-format=2.3]|}
\hline
{} & {APM-LR} & {Uncertainty} & {BALD} & {InfoGain} & {Random} & {MaxVar} \\  
\hline
{\emph{vehicle-full}} & 0.173 & 0.077 & 2.212 & 7.166 & 0.003 & 0.061 \\  
\hline
{\emph{vehicle-cars}} & 0.089 & 0.039 & 1.078 & 3.462 & 0.002 & 0.030 \\  
\hline
{\emph{vehicle-transport}} & 0.087 & 0.037 & 1.036 & 3.306 & 0.002 & 0.029 \\  
\hline
{\emph{letterDP}} & 0.336 & 0.149 & 4.230 & 12.755 & 0.005 & 0.118 \\  
\hline
{\emph{letterEF}} & 0.318 & 0.143 & 4.044 & 12.188 & 0.005 & 0.113 \\  
\hline
{\emph{letterIJ}} & 0.314 & 0.139 & 3.941 & 11.879 & 0.004 & 0.110 \\  
\hline
{\emph{letterMN}} & 0.331 & 0.147 & 4.170 & 12.531 & 0.005 & 0.117 \\  
\hline
{\emph{letterUV}} & 0.330 & 0.145 & 4.129 & 12.429 & 0.005 & 0.115 \\  
\hline
{\emph{letterVY}} & 0.318 & 0.143 & 4.063 & 12.284 & 0.004 & 0.114 \\  
\hline
{\emph{austra}} & 0.150 & 0.063 & 1.770 & 5.089 & 0.003 & 0.050 \\  
\hline
{\emph{wdbc}} & 0.125 & 0.052 & 1.480 & 6.415 & 0.003 & 0.042 \\  
\hline
{\emph{clouds}} & 0.119 & 0.053 & 1.522 & 2.735 & 0.002 & 0.041 \\  
\hline
{\emph{cross}} & 0.125 & 0.053 & 1.521 & 2.722 & 0.002 & 0.040 \\  
\hline
{\emph{horseshoe}} & 0.116 & 0.053 & 1.517 & 2.731 & 0.002 & 0.040 \\  
\hline
\end{tabular}
\end{small}

	\caption{\textbf{Cumulative selection time:} comparison of median cumulative time (s) for each method to select the first 40 examples (excluding seed points).}
	\label{tab:selection_cost}
\end{table}

Table \ref{tab:post_cost} isolates the compute time needed for performing VariationalEM at each input, summed over the first 40 examples. Interestingly, methods which are primarily focused on data space exploration (MaxVar, Random) require more time for variational posterior updating than exploitation methods (Uncertainty). Since VariationalEM is an iterative procedure that we run with an adaptive stopping rule (with convergence defined as the relative variational parameter difference falling below $1\mathrm{e}{-6}$ between iterations), it presumably requires more iterations to adjust to significant changes in the posterior distribution due to variability in examples. Although less accurate of an approximation than VariationalEM, using a Laplace posterior approximation instead would have a constant update time per method \citep{Jaakkola2000}.
\begin{table}[h]
	\centering
	\begin{small}
\begin{tabular}{|S[table-format=2.3]|S[table-format=2.3]|S[table-format=2.3]|S[table-format=2.3]|S[table-format=2.3]|S[table-format=2.3]|S[table-format=2.3]|}
\hline
{} & {APM-LR} & {Uncertainty} & {BALD} & {InfoGain} & {Random} & {MaxVar} \\  
\hline
{\emph{vehicle-full}} & 10.088 & 4.540 & 10.118 & 9.729 & 7.469 & 18.064 \\  
\hline
{\emph{vehicle-cars}} & 5.420 & 4.412 & 5.605 & 5.475 & 3.280 & 4.558 \\  
\hline
{\emph{vehicle-transport}} & 9.609 & 5.814 & 9.289 & 9.083 & 11.216 & 21.058 \\  
\hline
{\emph{letterDP}} & 7.618 & 6.412 & 6.904 & 6.758 & 10.694 & 11.851 \\  
\hline
{\emph{letterEF}} & 6.866 & 5.701 & 6.320 & 6.160 & 11.302 & 10.755 \\  
\hline
{\emph{letterIJ}} & 7.367 & 5.724 & 6.924 & 6.708 & 10.019 & 9.846 \\  
\hline
{\emph{letterMN}} & 8.190 & 6.281 & 7.615 & 7.375 & 10.082 & 13.236 \\  
\hline
{\emph{letterUV}} & 8.029 & 6.556 & 7.137 & 7.075 & 10.746 & 12.585 \\  
\hline
{\emph{letterVY}} & 7.463 & 5.760 & 7.142 & 6.910 & 8.234 & 9.975 \\  
\hline
{\emph{austra}} & 12.513 & 6.451 & 12.009 & 11.645 & 8.541 & 13.580 \\  
\hline
{\emph{wdbc}} & 17.966 & 10.880 & 14.183 & 13.874 & 20.763 & 29.778 \\  
\hline
{\emph{clouds}} & 1.221 & 1.172 & 1.156 & 1.322 & 3.201 & 5.318 \\  
\hline
{\emph{cross}} & 1.386 & 2.417 & 1.474 & 1.537 & 3.138 & 4.453 \\  
\hline
{\emph{horseshoe}} & 0.996 & 0.908 & 0.863 & 0.931 & 0.802 & 1.208 \\  
\hline
\end{tabular}
\end{small}

	\caption{\textbf{Cumulative VariationalEM time:} comparison of median cumulative time (s) for each method to perform VariationalEM over the first 40 examples (excluding seed points).}
	\label{tab:post_cost}
\end{table}

Table \ref{tab:all_cost} depicts the total compute time needed for selecting each example, performing VariationalEM, and retraining the logistic regression classifier at each iteration, summed over the first 40 examples. The median time needed for retraining the logistic regression classifier lies within 0.01 to 0.03 seconds across all methods and datasets, and therefore contributes only marginally to the total. While the spread of running times is more narrow than it would be when only evaluating selection time, the same general trend holds that InfoGain is more expensive than BALD and APM-LR.
\begin{table}[h]
	\centering
	\begin{small}
\begin{tabular}{|S[table-format=2.3]|S[table-format=2.3]|S[table-format=2.3]|S[table-format=2.3]|S[table-format=2.3]|S[table-format=2.3]|S[table-format=2.3]|}
\hline
{} & {APM-LR} & {Uncertainty} & {BALD} & {InfoGain} & {Random} & {MaxVar} \\  
\hline
{\emph{vehicle-full}} & 10.288 & 4.637 & 12.365 & 16.943 & 7.493 & 18.148 \\  
\hline
{\emph{vehicle-cars}} & 5.532 & 4.474 & 6.727 & 8.980 & 3.306 & 4.616 \\  
\hline
{\emph{vehicle-transport}} & 9.721 & 5.876 & 10.341 & 12.419 & 11.238 & 21.116 \\  
\hline
{\emph{letterDP}} & 7.992 & 6.583 & 11.139 & 19.534 & 10.730 & 11.995 \\  
\hline
{\emph{letterEF}} & 7.215 & 5.868 & 10.396 & 18.414 & 11.330 & 10.887 \\  
\hline
{\emph{letterIJ}} & 7.716 & 5.892 & 10.896 & 18.619 & 10.048 & 9.981 \\  
\hline
{\emph{letterMN}} & 8.561 & 6.455 & 11.813 & 19.991 & 10.124 & 13.374 \\  
\hline
{\emph{letterUV}} & 8.399 & 6.724 & 11.294 & 19.552 & 10.781 & 12.724 \\  
\hline
{\emph{letterVY}} & 7.802 & 5.931 & 11.233 & 19.233 & 8.260 & 10.118 \\  
\hline
{\emph{austra}} & 12.690 & 6.538 & 13.801 & 16.804 & 8.574 & 13.655 \\  
\hline
{\emph{wdbc}} & 18.130 & 10.968 & 15.711 & 20.323 & 20.787 & 29.842 \\  
\hline
{\emph{clouds}} & 1.358 & 1.241 & 2.706 & 4.122 & 3.224 & 5.385 \\  
\hline
{\emph{cross}} & 1.534 & 2.490 & 3.028 & 4.291 & 3.159 & 4.515 \\  
\hline
{\emph{horseshoe}} & 1.134 & 0.978 & 2.405 & 3.741 & 0.819 & 1.264 \\  
\hline
\end{tabular}
\end{small}

	\caption{\textbf{Cumulative running time:} comparison of median cumulative run time (s) for each method to select each example, perform VariationalEM, and retrain the logistic regression classifier over the first 40 examples (excluding seed points).}
	\label{tab:all_cost}
\end{table}

\subsection{Failure Mode Analysis}
\label{app:fail}
While in many cases APM-LR performs comparably to InfoGain, BALD, and Uncertainty while outperforming Random and MaxVar, the main exception in our experiments is on \emph{vehicle-cars} (Figure \ref{fig:vehicle-cars}), where APM-LR, Random, and MaxVar outperform InfoGain, BALD, and Uncertainty. Conceptually, what differentiates these two classes of methods is that APM-LR, Random, and MaxVar have explicit exploration components to their selection policies, while InfoGain, BALD, and Uncertainty only seek to directly maximize information or uncertainty. As we will demonstrate below, on \emph{vehicle-cars} this difference in exploration correlates with significant differences in generalization performance.

To isolate the effect of each term in APM-LR (eq.\ \eqref{eq:apmlr-supp}) --- corresponding to exploitation and exploration --- we simulated two pseudo-APM policies where only one of the terms is active at once. In \emph{APM-LR-U}, examples are selected that minimize the first term, which has an action similar to uncertainty sampling:
\begin{align*}
\text{\emph{APM-LR-U}:}\quad x_n &= \argmin_{x \in \Ucal_n}\enspace(\mu_n^T x)^2.
\intertext{In \emph{APM-LR-V}, examples are selected that minimize the second term, which prefers examples that probe in directions of high posterior variance:}
\text{\emph{APM-LR-V}:}\quad x_n &= \argmin_{x \in \Ucal_n}\enspace\biggl(\sqrt{x^T \Sigma_n x}-\sqrt{\frac{2}{\pi}P_n}\biggr)^2.
\end{align*}

We start in Figure \ref{fig:vehicle-cars-all} by plotting generalization performance as in Figure \ref{fig:vehicle-cars}, with the addition of APM-LR-U and APM-LR-V. In all plots below, error bars are removed for visual clarity, and the query horizon spans the entire training sequence (until the training pool is exhausted). As expected, APM-LR-V performs comparably to MaxVar, since both methods prefer examples that probe in directions of large posterior variance. Similarly, APM-LR-U performs comparably to Uncertainty, since both methods minimize distance to a hyperplane estimate (the former using the posterior mean hyperplane, the latter using a MAP estimate). These results support the hypothesis that it is the exploration component of APM-LR which leads to improved performance on \emph{vehicle-cars} over non-exploration methods, including its own exploitation variant APM-LR-U.

\def\fh{2.5in}
\begin{figure}[h]
	\centering
	\includegraphics[height=\fh]{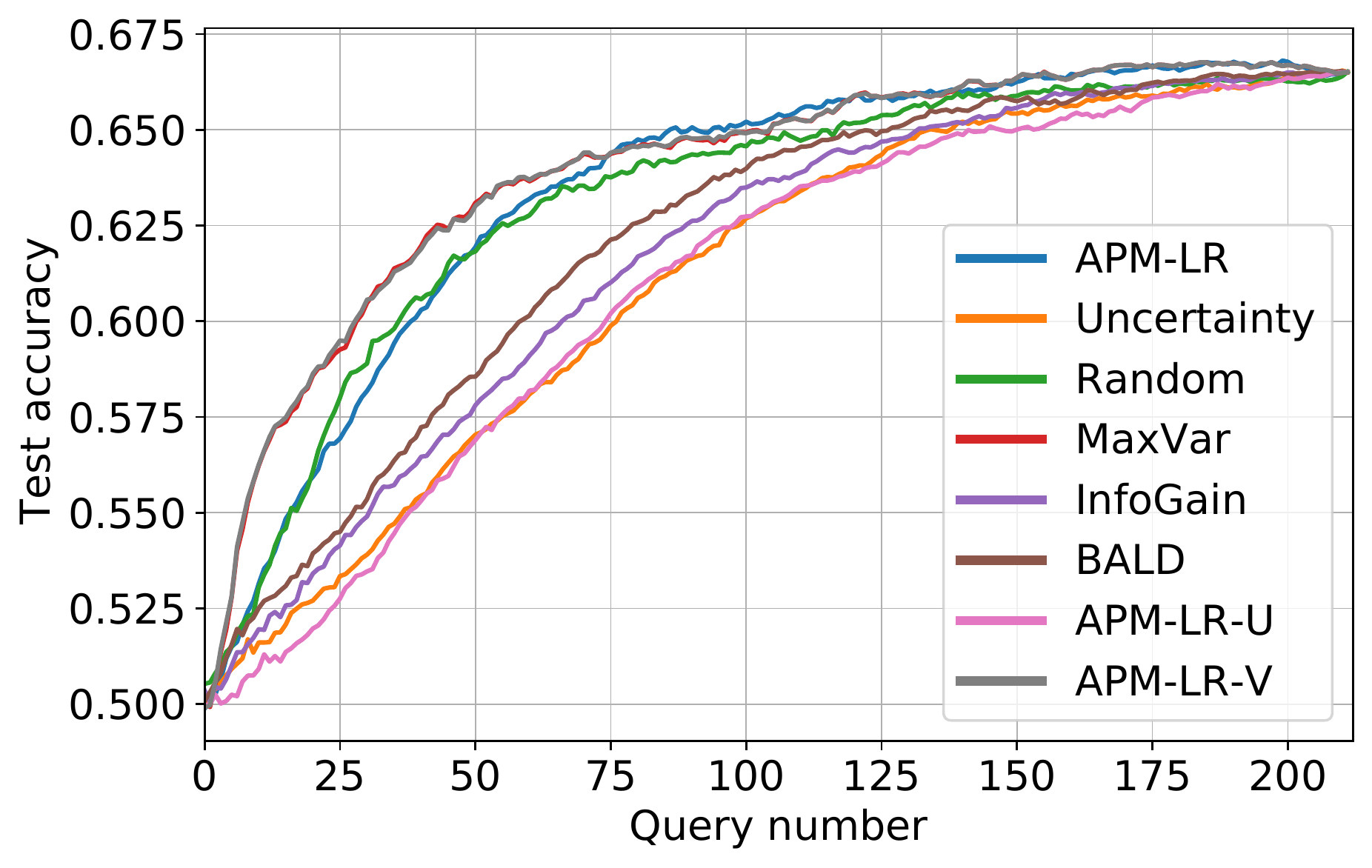}
	\caption{Test accuracy on \emph{vehicle-cars}, over expanded method set.}
	\label{fig:vehicle-cars-all}
\end{figure}

\begin{figure}[h]
	\centering
	\includegraphics[height=\fh]{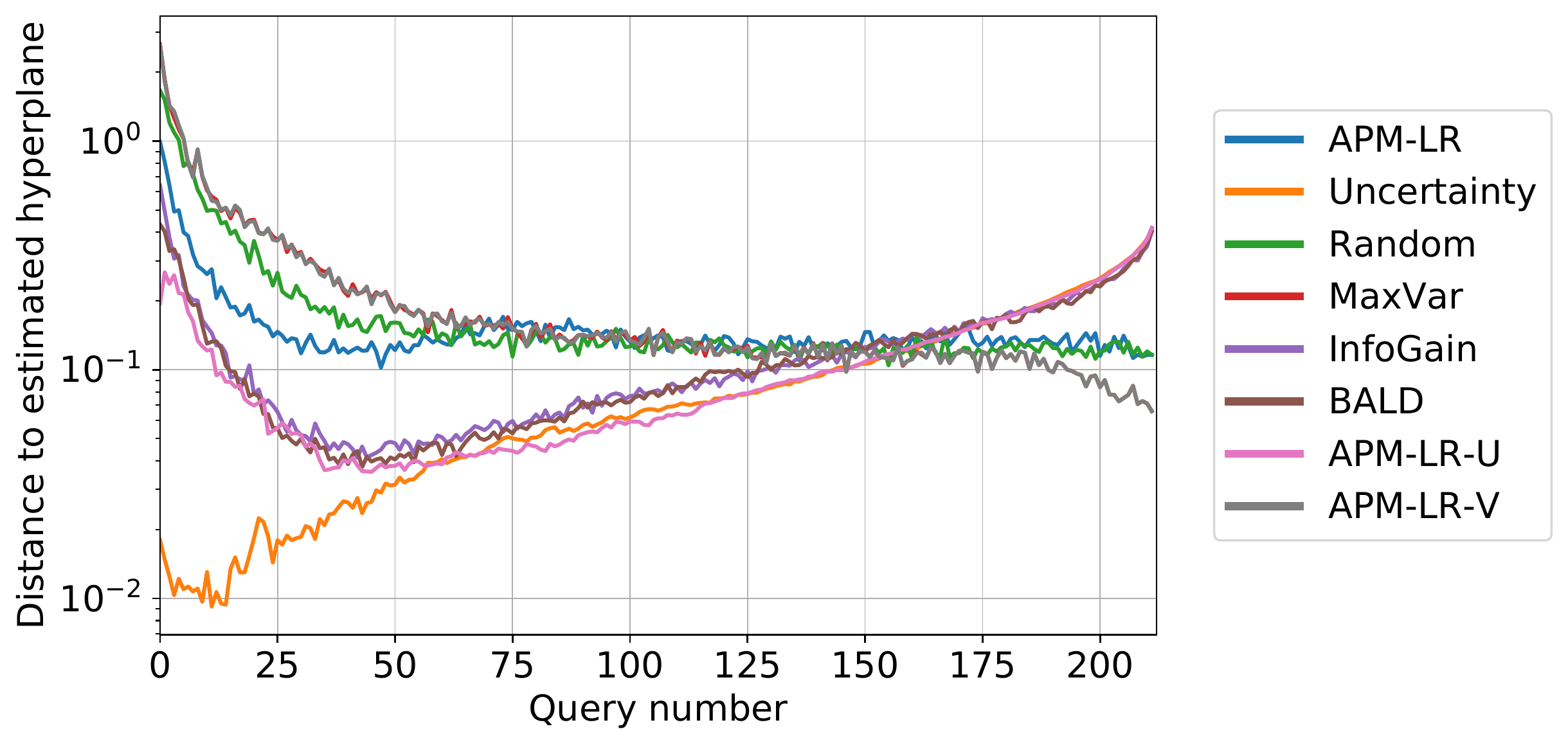}
	\caption{Exploitation metric for \emph{vehicle-cars}: average distance of selected example to estimated hyperplane. Small distances reflect high levels of policy exploitation since this reflects examples being queried that are uncertain with respect to the current hyperplane estimate.}
	\label{fig:vehicle-cars-hypdist}
\end{figure}

We can explore this hypothesis further by directly evaluating metrics for exploitation and exploration of each method. To measure exploitation, in Figure \ref{fig:vehicle-cars-hypdist}, we plot the average distance from each selected example to the MAP hyperplane estimate. Since distance from the classifier hyperplane directly corresponds to label uncertainty in logistic regression, this distance is a direct measure of how often a policy selects uncertain examples. By definition, Uncertainty begins by querying examples that are closest to the hyperplane estimate, maximally exploiting the estimate to query examples with the highest model uncertainty. The remaining methods vary in their levels of initial distance from the hyperplane estimate, but all eventually query close to their respective estimates, either by design or due to exhausting the full training pool. Notably, the level of initial distance from the hyperplane corresponds almost exactly to test accuracy performance: high-performing MaxVar and APM-LR-V initially query far from their hyperplane estimates, while the poorly performing Uncertainty queries examples close by.

\begin{figure}[h]
	\centering
	\begin{subfigure}[t]{0.5\textwidth}
		\centering
		\includegraphics[width=\textwidth]{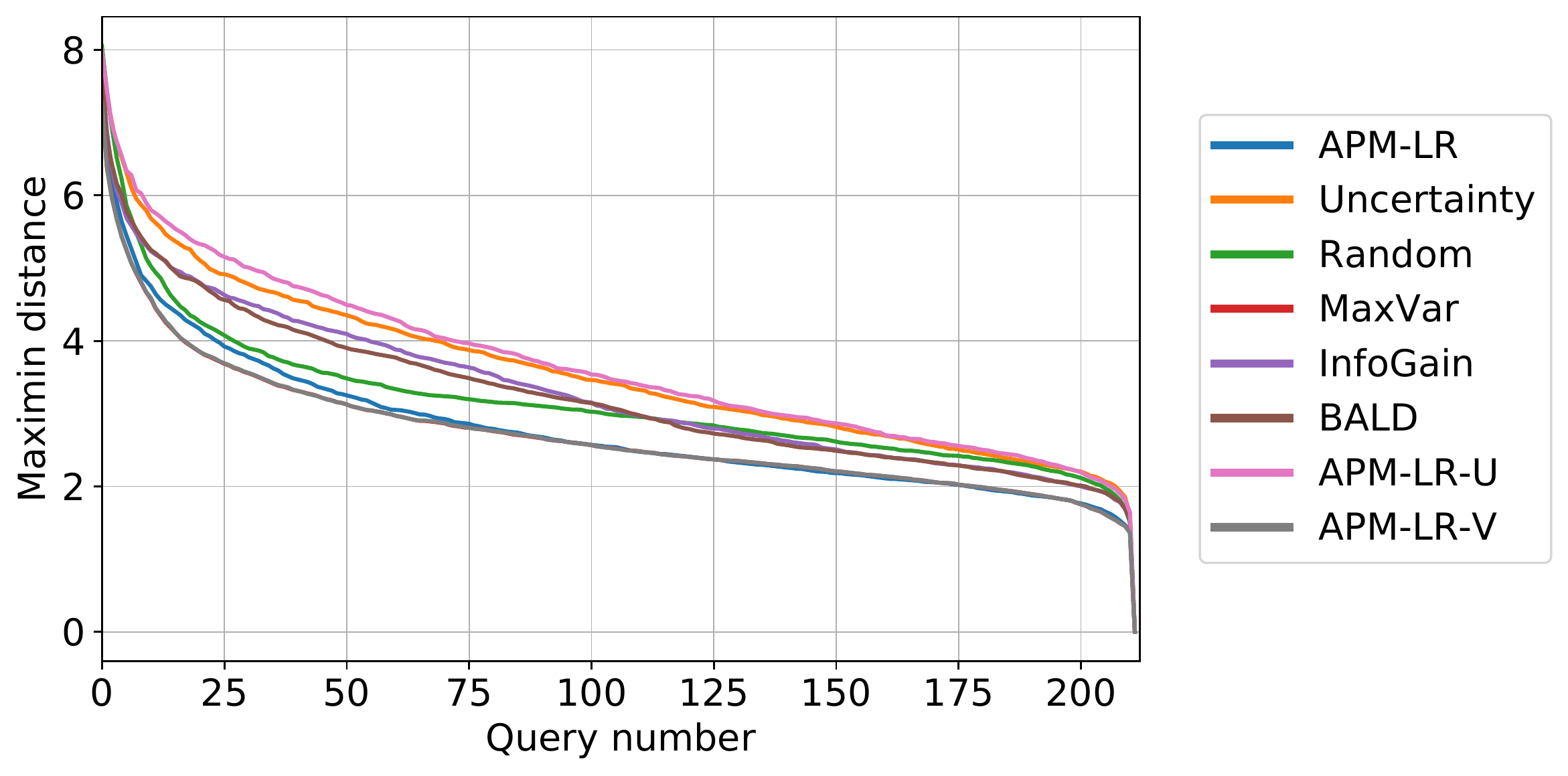}
		\caption{}
		\label{fig:vehicle-cars-maxmin}
	\end{subfigure}%
	~
	\begin{subfigure}[t]{0.5\textwidth}
		\centering
		\includegraphics[width=\textwidth]{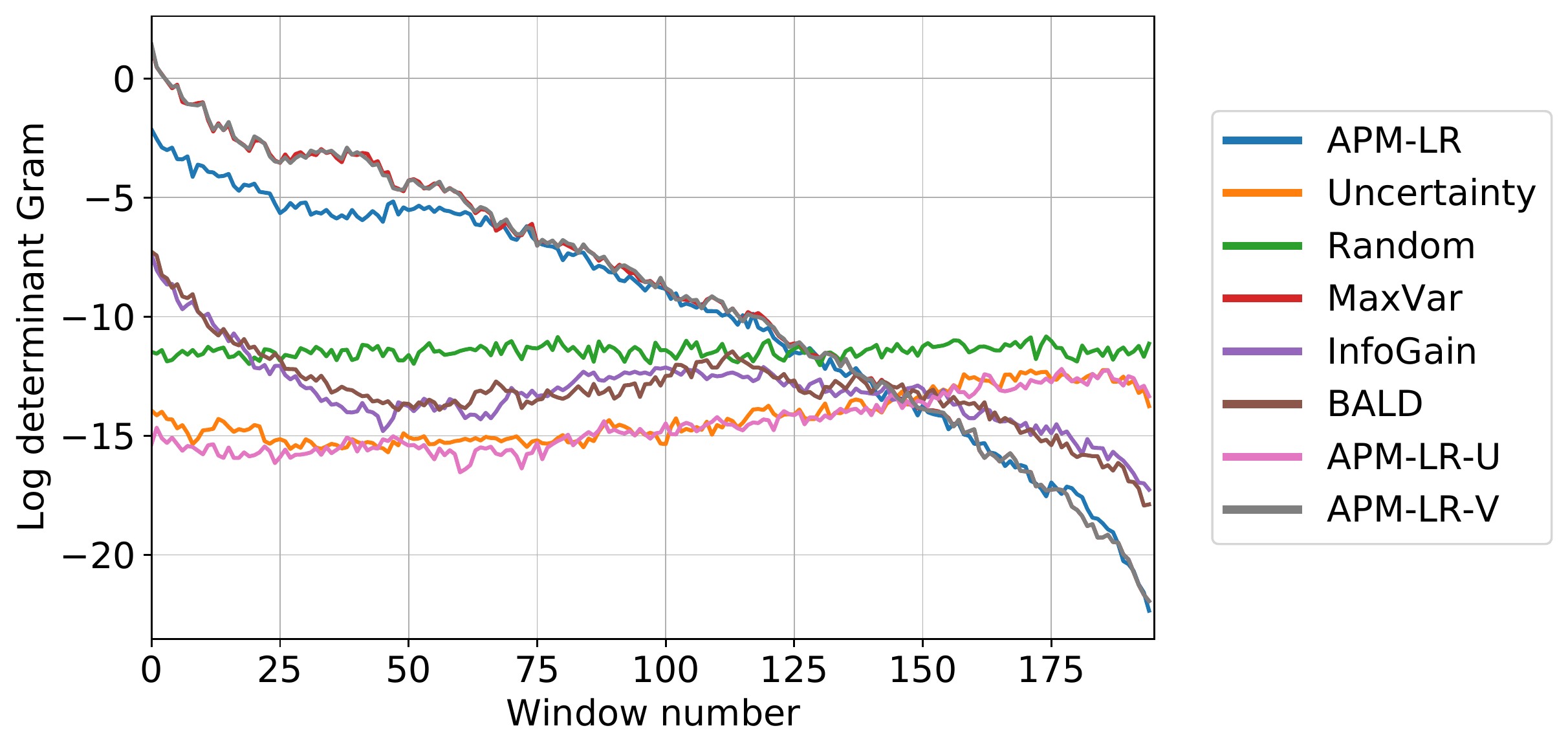}
		\caption{}
		\label{fig:vehicle-cars-gram}
	\end{subfigure}%
	\caption{Exploration metrics for \emph{vehicle-cars}: (a) maximum distance from an unlabeled example to its closest labeled example. Smaller values indicate lower levels of unlabeled data isolation, and correspond to higher levels of exploration. (b) Log determinant of Gram matrix, where larger values correspond to higher levels of exploration.}
	\label{fig:vehicle-cars-explore}
\end{figure}

To measure policy exploration, we use two metrics and plot their average values in Figure \ref{fig:vehicle-cars-explore}. In the first metric, we measure the Euclidean distance from each unlabeled example to its nearest labeled neighbor, and take the maximum such distance over all unlabeled examples. This quantity measures the worst-case level of isolation of an unlabeled point to its nearest labeled neighbor, with lower values corresponding to higher degrees of policy exploration. A similar quantity is involved in the construction of coresets for active learning to promote diversity among selected examples \citep{sener2018active}. As our second metric, we consider windows of $d$ examples (recall that $d$ denotes the data space dimension) and plot the log determinant of the Gram matrix of the examples selected in each window, which can be used as a measure of example diversity (higher values correspond to higher levels of example diversity) \citep{Ash2020Deep}. In Figure \ref{fig:vehicle-cars-maxmin}, MaxVar, APM-LR-V, APM-LR, and Random have the lowest average maximin distances, corresponding to lower levels of isolated unlabeled examples. Similarly, these methods generally have large initial Gram matrix log determinants, as depicted in Figure \ref{fig:vehicle-cars-gram}.

The ablation of individual terms in APM-LR and direct measurement of exploitation and exploration of each active learning method suggests that when tested on \emph{vehicle-cars}, exploration-based methods outperform methods that do not explicitly optimize for diverse selection. While this extended analysis is limited to a single dataset, it provides evidence that the exploration term in APM-LR can lead to higher levels of performance on a real-world dataset, where methods that do not directly account for exploration might fail.

\end{document}